%% file: acl2023.tex
\pdfoutput=1

\documentclass[11pt]{article}

\usepackage{ACL2023}

\usepackage{times}
\usepackage{latexsym}

\usepackage{multirow}
\usepackage{multibib}
\usepackage{booktabs}
\usepackage{color}
\usepackage{graphicx}
\usepackage{subfigure}
\usepackage{tikz}
\usepackage{pgfplots}
\usepackage{caption}
\usepackage{amsmath}
\usepackage{makecell}
\usepackage{hyperref}
\usepackage{arydshln}
\usepackage{bm}
\usepackage{array}
\usepackage{amsfonts}
\usepackage{times}
\usepackage{latexsym}
\usepackage{color}
\usepackage{tikz}
\usepackage{pgfplots}
\pgfplotsset{compat=1.18}
\usepackage{graphicx}
\usepackage{caption}
\usepackage{amsmath}
\usepackage{makecell}
\usepackage{hyperref}
\usepackage{amssymb}

\usepackage{CJKutf8}

\usepgfplotslibrary{groupplots} 
\usepgfplotslibrary[groupplots]
\usetikzlibrary{pgfplots.groupplots}
\usetikzlibrary[pgfplots.groupplots]

\usetikzlibrary{backgrounds}
\usetikzlibrary{positioning,fit,calc}
\usetikzlibrary{shadows}

\usepackage{xcolor}
\usepackage{colortbl}
\definecolor{ugreen}{rgb}{0,0.5,0}
\definecolor{myblue}{RGB}{30,144,255}
\definecolor{myyellow}{RGB}{255,235,205}
\definecolor{myred}{RGB}{255,99,71}
\definecolor{mypurple}{RGB}{238,229,248}
\definecolor{mygreen}{RGB}{202,235,216}

\definecolor{dblue}{RGB}{19,103,131}
\definecolor{dred}{RGB}{200,36,35}
\definecolor{lpurple}{RGB}{137,131,191}
\definecolor{pink}{RGB}{230,25,75}
\definecolor{dpink}{RGB}{199,109,162}
\definecolor{lpink}{RGB}{229,123,127}
\definecolor{dpurple}{RGB}{145,30,180}
\definecolor{brown}{RGB}{109,1,31}
\definecolor{dgrey}{RGB}{60,64,91}
\definecolor{dgreen}{RGB}{14,96,107}
\definecolor{lgreen}{RGB}{127,203,164}
\definecolor{blgreen}{RGB}{31,146,139}
\definecolor{dyellow}{RGB}{187,151,39}
\definecolor{lyellow}{RGB}{255,195,0}
\definecolor{blpurple}{RGB}{142,139,254}
\definecolor{orange}{RGB}{250,109,29}
\definecolor{lorange}{RGB}{252,154,107}

\usepackage[T1]{fontenc}

\usepackage[utf8]{inputenc}

\usepackage{microtype}
\usepackage{pifont}
\usepackage{inconsolata}
\renewcommand{\thefootnote}{\fnsymbol{footnote}}

\title{CTC-based Non-autoregressive Speech Translation}

\author{Chen Xu\textsuperscript{1}\footnotemark[2] , Xiaoqian Liu\textsuperscript{1}, Xiaowen Liu\textsuperscript{1}, Qingxuan Sun\textsuperscript{1}, {\bf Yuhao Zhang\textsuperscript{1}}, \\
{\bf Murun Yang\textsuperscript{1},} 
{\bf Qianqian Dong\textsuperscript{2},} {\bf Tom Ko\textsuperscript{2},} {\bf Mingxuan Wang\textsuperscript{2}\footnotemark[1] ,}\\
{\bf Tong Xiao\textsuperscript{1,3}\footnotemark[1] ,} {\bf Anxiang Ma\textsuperscript{1,3},} {\bf Jingbo Zhu\textsuperscript{1,3}} \\
  \textsuperscript{1}School of Computer Science and Engineering, Northeastern University, Shenyang, China\\
  \textsuperscript{2}ByteDance \\
  \textsuperscript{3}NiuTrans Research, Shenyang, China \\
  \texttt{\{xuchennlp, liuxiaoqian0319, liuxiaowenneu\}@outlook.com} \\
  \texttt{\{dongqianqian, tom.ko, wangmingxuan.89\}@bytedance.com}\\
  \texttt{\{xiaotong, maanxiang, zhujingbo\}@mail.neu.edu.cn} 
  }

\begin{document}

\begin{CJK*}{UTF8}{gbsn}

\maketitle
\begin{abstract}

\footnotetext[1]{Corresponding author.}
\footnotetext[2]{Work was done while at ByteDance AI Lab.}
\renewcommand{\thefootnote}{\arabic{footnote}}
Combining end-to-end speech translation (ST) and non-autoregressive (NAR) generation is promising in language and speech processing for their advantages of less error propagation and low latency.
In this paper, we investigate the potential of connectionist temporal classification (CTC) for non-autoregressive speech translation (NAST).
In particular, we develop a model consisting of two encoders that are guided by CTC to predict the source and target texts, respectively.
Introducing CTC into NAST on both language sides has obvious challenges: 1) the conditional independent generation somewhat breaks the interdependency among tokens, and 2) the monotonic alignment assumption in standard CTC does not hold in translation tasks.
In response, we develop a prediction-aware encoding approach and a cross-layer attention approach to address these issues.
We also use curriculum learning to improve convergence of training.
Experiments on the MuST-C ST benchmarks show that our NAST model achieves an average BLEU score of 29.5 with a speed-up of 5.67$\times$, which is comparable to the autoregressive counterpart and even outperforms the previous best result of 0.9 BLEU points\footnote{The code is available at \href{https://github.com/xuchennlp/S2T}{https://github.com/xuchennlp/\\S2T.}}.

\end{abstract}

\section{Introduction}

End-to-end speech translation (E2E ST) has attracted unprecedented attention and achieved dramatic development in recent years \cite{Duong_naacl2016, Berard_arxiv2016, Weiss_ISCA2017, Anastasopoulos_NAACL2018, Wang_aaai2020, Wang_acl2020, Xu_ACL2021, zhang2022improving}.
Stand-alone modeling reduces the inference latency by almost half compared to cascaded systems, where the automatic speech recognition (ASR) model and the machine translation (MT) model run serially.
This helps the application in real scenarios, especially with limited computational resources.

However, this advantage only holds in the context of autoregressive (AR) decoding, where each token is generated depending on the previously predicted results.
Non-autoregressive (NAR) generation \cite{Gu_ICLR2018}, the recently popular decoding method in ASR and MT, makes the inference process fast by predicting the output sequence in parallel, resulting in the E2E ST no longer being superior in terms of inference speed-up.

A natural question arises: can we build a powerful non-autoregressive speech translation (NAST) model?
The NAR results in the latest literature are still inferior to the AR counterparts with a large gap of about 2 $\sim$ 3 BLEU points, even with the iterative refinement process \cite{Inaguma_CORR2021}.
In this work, we aim to develop a promising NAST model for comparable performance to the AR model without complex decoding.

We resort to the connectionist temporal classification (CTC, \citealp{Graves_ACL2006}) because of its great success in ASR and MT and the convenience of variable length prediction.
CTC is well suited for speech-to-text modeling, where the input sequence is longer than the output.
Recent studies show that CTC-based NAR models achieve comparable or even better performance than their AR counterparts, providing insight into the design of the powerful CTC-NAST model.

Our CTC-NAST model is decoder-free and consists of two stacked encoders: an acoustic encoder and a textual encoder.
They are guided by CTC to predict transcription and translation, respectively \cite{Chuang_ACL2021}.
Then, we carry out a careful and systematic inspection of the underlying issues and address the challenges of CTC-NAST. 
In particular,
\begin{itemize}
    \item The conditional independence assumption allows fast inference but omits interdependency across the whole sequence.
    We identify the \textit{prediction-aware encoding} (PAE) method underlying the success of a series of studies \cite{Nozaki_ISCA2021, Huang_AAAI2022, Higuchi_ASRU2021}, which observe preliminary prediction and refine it in the final generation.
    Following this idea, we predict the CTC result in the intermediate layer and then integrate it into the subsequent encoding.
    
    \item Another inherent property of CTC, the monotonic assumption, is valid for ASR but does not hold for translation tasks, where a future word in the target text may be aligned with the earlier part of the source text, especially on distant language pairs \cite{hannun2017distill}.
    A critical requirement of the decoder-free design is the \textit{reordering augmentation} \cite{Chuang_ACL2021}. 
    As a remedy, we introduce an additional cross-layer attention module, which is complementary to the self-attention module.	
\end{itemize}

Even with the above efforts, NAST is still a difficult task that suffers from heavy modeling burdens.
A \textit{curriculum learning strategy} that guides the training in an easy-to-hard way is significant for better convergence.
We replace part of the incorrect prediction with ground truth in PAE to prompt the generation of the whole sequence.  
In this way, the model relieves the CTC learning burden by observing almost the whole sequence in the early stages, while only a few tokens are replaced as CTC performance improves, ensuring consistency between training and inference.

Our CTC-NAST model is simple, completely parallel, and works well for both similar and distant language pairs.
The proposed methods yield a remarkable gain of 3.0 BLEU points on MuST-C En-De, achieving an average BLEU score of 29.5 with an inference speed-up of 5.67$\times$, and even outperforming the best previous AR results by 0.9 BLEU points.
We also report competitive results on the more challenging MuST-C En-Ja and Fisher-Callhome corpus.

\section{Background}

\subsection{Connectionist Temporal Classification}


CTC \cite{Graves_ACL2006} was originally proposed for labeling unsegmented sequences.
It learns monotonic alignment between acoustic features and transcriptions, which is valid for cross-modal learning like ASR.
CTC helps convergence and allows re-scoring decoding through a lightweight output layer, achieving great success in ASR as an auxiliary loss on top of the encoder \cite{Watanabe_IEEE2017, Karita_ISCA2019}.
Given the encoder representation $h$ and the corresponding sequence $y$, the CTC loss is defined as:
\begin{eqnarray}
    \mathcal{L}_{\rm CTC} = -{\rm log} \textrm{P}_{\rm CTC}(y|h)
\end{eqnarray}
where the probability is calculated by marginalizing over all possible alignments $\Phi(y)$ between $h$ and $y$:
\begin{eqnarray}
    \textrm{P}_{\rm CTC}(y|h) = \sum_{\pi \in \Phi(y)} \textrm{P}(\pi | h)
\end{eqnarray}

CTC has the same conditional independence property as NAR generation, where the probability of the path $\pi$ is the product of the probability $P(\pi_t|h_t)$ at each time step $t$:
\begin{eqnarray}
    \textrm{P}(Y|X) \approx \prod_{t=1}^{T} \textrm{P}(\pi_{t} | h_t)
\label{ctc_prob}
\end{eqnarray}
where $T$ is the length of $h$.


\subsection{AR and NAR}



Given a source sequence $X = (x_1, \cdots, x_{T^{'}})$, a sequence-to-sequence model predicts the target sequence $Y = (y_1, \cdots, y_T)$ by conditional distribution:
\begin{eqnarray}
    \textrm{P}(Y|X; \theta) = \prod_{t=1}^{T} \textrm{P}_{\textrm{AR}}(y_t|y_{<t}, X; \theta)
    \label{ar_prob}
\end{eqnarray}
where $\theta$ is the model parameters. 
This autoregressive generation learns sequential dependency but suffers from high inference latency.




Instead, NAR carries out the conditional independent prediction for parallel inference \cite{Gu_ICLR2018}:
\begin{eqnarray}
    \textrm{P}(Y|X; \theta) = \prod_{t=1}^{T} \textrm{P}_{\textrm{NAR}}(y_t| X; \theta)
    \label{nar_prob}
\end{eqnarray}
Although the vanilla NAR model speeds up inference by about $15 \times$ \cite{Gu_ICLR2018}, it is still inferior to the AR counterpart by a large gap.

Researchers have proposed many series of methods to improve the generation quality and investigate a better trade-off between performance and speed in the MT task, such as the iterative decoding method \cite {Lee_EMNLP2018, Stern_ICML2019, Ghazvininejad_EMNLP2019, Kasai_ICML2020}, latent variable method \cite{Gu_ICLR2018, Song_EMNLP2021, Gu_ACL2021}, data manipulation method \cite {Zhou_ACL2020, Bao_ACL2022, Ding_ACL2020}, enhancement based method \cite{Guo_AAAI2019, Wang_AAAI2019}, and semi-autoregressive decoding \cite{Ran_ACL2020}.
There are also some studies to design the architecture of the NAR models, such as the use of CTC for prediction for its ability of variable length prediction \cite{Libovick_EMNLP2018, Shu_AAAI2020, Saharia_EMNLP2020}. 

In addition, the NAR generation also shows promising results in ASR task, especially the CTC-based systems \cite{higuchi20b_interspeech, Higuchi_ICASSP2021, Lee_ICASSP2021, Nozaki_ISCA2021, Kim_Corr2022}.


\subsection{Speech Translation}

Recently, E2E ST has received a lot of attention due to its direct modeling \cite{Berard_arxiv2016}.
Unlike the conventional cascaded system that decouples the cross-modal and cross-lingual modeling into ASR and MT models respectively \cite{Ney_IEEE1999, Mathias_ICASSP2006}, the end-to-end manner is more elegant and has the potential for fast inference and error-free propagation.


One promising route to improve ST is to develop more adaptive architectures according to the task characteristics.
Based on the idea of modeling decoupling, the stacked encoding method divides cross-modal and cross-lingual learning into acoustic and semantic encoders, respectively \cite{Liu_corr2020, Xu_ACL2021}.
In this design, the CTC loss for transcription is usually introduced to guide the learning of the acoustic encoder, which significantly helps convergence.
In addition, the latent alignment learned in the CTC is used to bridge the two encoders.
\newcite{Liu_corr2020} shrink the sequence length based on CTC prediction.
\newcite{Xu_ACL2021} introduce an adapter to bridge two encoders by integrating CTC prediction.

Several studies investigate the NAR generation in ST \cite{Inaguma_CORR2021, Inaguma_ICASSP2021, Chuang_ACL2021}.
However, current NAR systems are still inferior to AR counterparts, especially CTC-based systems.
Researchers also continue to extend the use of CTC to learn target text as an auxiliary loss of the encoder \cite{Zhang_corr2022, Yan_Corr2022}.
But there is no work to inspect the underlying issues in the CTC modeling of target text in ST.
To this end, we study the challenges of building a powerful CTC-based NAST model and then propose corresponding methods.
We also extend our method to AR models for a comprehensive exploration.

\input{fig/fig1.tex}

\section{CTC-NAST}

Among many well-established NAR designs for ASR or MT models, CTC is particularly suitable for ST modeling because the input length is remarkably longer than its output.
In this section, we present CTC-NAST in detail.
We first describe the base architecture, then identify and address three underlying challenges.
See Figure \ref{fig:fig1} for an overview of our system.

\subsection{Base Architecture}

ST aims to translate audio in the source language to text in the target language directly.
Let $(x; y^s; y^t)$ be a training sample of ST, where $x$ is the input speech feature sequence, $y^s$ is the corresponding transcription of $x$, and $y^t$ is the translation in the target language.
We assume that transcription is always available in our work.

We drop the decoder network and rely only on the CTC-based encoder.
Following the design of SATE \cite{Xu_ACL2021, Chuang_ACL2021}, we decouple the encoding into an acoustic encoder and a textual encoder in a stack architecture, as shown in Figure \ref{fig:a}.
They are guided by CTC loss for transcription and translation (denoted CTC and XCTC for distinction), respectively.

Formally, given a representation $h^a$ of the acoustic encoder output, the CTC loss is calculated as:
\begin{eqnarray}
    \mathcal{L}_{\rm CTC} = -{\rm log} \textrm{P}_{\rm CTC}(y^s|h^a)
\end{eqnarray}
Similarly, the XCTC loss is calculated as:
\begin{eqnarray}
    \mathcal{L}_{\rm XCTC} = -{\rm log} \textrm{P}_{\rm XCTC}(y^t|h^t)
\end{eqnarray}
where $h^t$ is the representation of the textual encoder output.

Then, the training objective is formulated as the interpolation of the two CTC losses:
\begin{eqnarray}
    \mathcal{L} = \alpha_A \cdot \mathcal{L}_{\rm CTC} + \alpha_T \cdot \mathcal{L}_{\rm XCTC}
    \label{two_loss}
\end{eqnarray}
where $\alpha_{\rm A}$ and $\alpha_{\rm T}$ are the coefficients of the CTC and XCTC losses, respectively.

Although CTC works well for the NAR ASR model, extending CTC naively to the more challenging ST task is fragile.
We claim that CTC-NAST can be improved by addressing three issues:
\begin{itemize}
    \item \textbf{Conditional independence assumption} is an inherent property of CTC, which ignores interdependency with past or future contexts, leading to poor generation \cite{Chan_ICML2020}, like repetition and omission errors.
    \item Although the self-attention network has the modest reordering capability \cite{Chuang_ACL2021}, our encoder-only architecture is hard to handle the \textbf{monotonic assumption}, especially for distant language pairs.
    \item E2E ST already suffers from the heavy burden of cross-modal and cross-lingual mapping, while NAR modeling further aggravates the difficulty and results in \textbf{poor convergence}.
\end{itemize}



\subsection{Prediction-aware Encoding}


NAR generation enlarges the search space in inference due to conditional independence \cite{Ran_AAAI2021}, especially with the long speech sequence of hundreds and thousands of units. 
A commonly-used solution, incorporating latent variables that contain the initial prediction into modeling, has been demonstrated to be effective \cite{Lee_EMNLP2018}.
In this way, the NAR generation is decoupled as the multiple-step refinement of the target sequence, enabling the model to be aware of the previous prediction.


Inspired by the prior efforts in MT \cite{Huang_AAAI2022} and ASR \cite{Nozaki_ISCA2021}, we introduce prediction-aware encoding (PAE).
The detailed illustration is shown in Figure \ref{fig:c}.
Specifically, given one representation $h^l$ outputted by the intermediate encoder layer $l$, PAE integrates the prediction information (corresponding \textcircled{\textbf{\textsl{\small 1}}} in the Figure) into the following encoding explicitly by weighting the embedding matrix $W$ over the current CTC distribution (called InterCTC) \cite{Xu_ACL2021}:
\begin{eqnarray}
    \label{pae}
    \textrm{PAE}(h^l) = h^l + \textrm{P}_{\rm InterCTC}(\pi|h^l) \cdot W
\end{eqnarray}
where the weights $W$ are shared in the whole network. 
Note that we use PAE to augment the learning of both CTC and XCTC.

Since the poor prediction leads to the risk of error propagation, we also optimize the InterCTC loss for guaranteed prediction:
\begin{eqnarray}
    \mathcal{L_{\rm InterCTC}} = -{\rm log} \textrm{P}_{\rm InterCTC}(y|h)
\end{eqnarray}

In this way, we ensure that CTC predicts well. 
However, the worse result for XCTC limits the benefits of PAE, which may result in negative effects. We alleviate this issue in Section \ref{CL}.

Now, we re-formulate the training loss in Eq. \ref{two_loss} as:
\begin{eqnarray}
    \mathcal{L} & = & \alpha_{\rm A} \cdot \mathcal{L}_{\rm CTC} + \alpha_{\rm T} \cdot \mathcal{L}_{\rm XCTC} \nonumber \\
    & + & \beta_{\rm A} \cdot \frac{1}{M}\sum_{m=1}^{m} \mathcal{L}_{\rm InterCTC}^m \nonumber \\
    & + & \beta_{\rm T} \cdot \frac{1}{N}\sum_{n=1}^{N} \mathcal{L}_{\rm InterXCTC}^n
\end{eqnarray}
where $M$ and $N$ are the numbers of the intermediate CTC and XCTC, $\beta_{\rm A}$ and $\beta_{\rm T}$ are the corresponding coefficients.

\subsection{Reordering Augmentation}

Vanilla Transformer generates each token by distributing the weight of the encoder-decoder attention module to the corresponding source part to be translated, which easily handles the order gap between languages.
However, CTC modeling faces the intractable issue of reordering the representation into the target language order during encoding.
Although previous studies have demonstrated that the MT or ST encoder can capture the global information \cite{Yang_emnlp2018, Xu_ACL2021}, it is still difficult to rely only on the self-attention module to search the positions that contribute significantly to decoding \cite{Chuang_ACL2021}.

To enhance the reordering capability of CTC-NAST, we mimic the design of the decoder and introduce cross-layer attention (CLA) module, which is inserted between the self-attention module and the feed-forward module in the specific layers of the textual encoder, as shown in Figure \ref{fig:b}.
Let $\textrm{SA}(\cdot, \cdot, \cdot)$ and $\textrm{CLA}(\cdot, \cdot, \cdot)$ denote the self-attention and CLA modules, the new Transformer layer $j$ can be formulated as:
\begin{eqnarray}
    h^{'} & = & h^{j-1} + \textrm{SA}(h^{j-1}, h^{j-1}, h^{j-1}) \\
    h^{'} & = & h^{'} + \textrm{CLA}(h^{'}, h^k, h^k) \\
    h^{j} & = & h^{'} + \textrm{FFN}(h^{'})
\end{eqnarray}
where $h^k$ is the representation output from the layer $k (k < j)$.

In this way, CLA offers a remedy for the lacking attention, that captures the information from the bottom layer directly and is complementary to the self-attention module.
Now the textual encoder acts as both a stack of the encoder and the decoder of the vanilla encoder-decoder model.

In order to further enhance the ability of CLA, we introduce the drop-net technique. 
In each layer containing the CLA module, we drop the self-attention module with a probability $p_{drop} \in [0, 1]$.
Note that the self-attention module always keeps during inference.

\subsection{Curriculum Learning Strategy}
\label{CL}

Even with the auxiliary encoding and improved design architecture, the CTC-NAST model still faces the difficulty of a heavy modeling burden, leading to poor convergence.
Inspired by \newcite{Qian_ACL2021}, a curriculum learning strategy is remarkably important to reduce the dependency in the early stage and increase the difficulty along the training process.

As illustrated in Figure \ref{fig:c}, we replace part of the prediction (corresponding \textcircled{\textbf{\textsl{\small 1}}} in the Figure) in Eq. \ref{pae} with the ground truth (corresponding \textcircled{\textbf{\textsl{\small 2}}} in the Figure), which mitigates the negative effects of error propagation caused by the poor XCTC performance in PAE and prompts the generation of the whole sequence.
Unlike the same lengths between input and output in the decoder, the length of the input acoustic feature is remarkably longer than the corresponding text in CTC.
Therefore, we take the best alignment computed by the model as the ground truth \cite{Gu_ACL2021, Huang_AAAI2022}:
\begin{eqnarray}
    \hat{\pi} = \arg \max_{\pi \in \Phi(y)} \textrm{P}(\pi | s; \theta^{'})
\end{eqnarray}
where $\theta^{'}$ is the current model parameter. 
Note that the length of $\hat{\pi}$ is the same as the input.

Denote the replacement ratio as $r \in [0, 1]$, we uniformly sample a random variable $U$ from $[0, 1]$:
\begin{eqnarray}
     \hat{P_t} = \mathbb{I}(U >= r) * p_t + \mathbb{I}(U < r) * \hat{\pi}_t 
\end{eqnarray}
where $\mathbb{I}(\cdot)$ is the indicator function. 

\begin{table*}[t!]
  \centering
  \resizebox{\textwidth}{!}{
  \normalsize
  \begin{tabular}{clccccccccccc}
    \toprule
    & Model &  De & Es & Fr & It & Nl & Pt & Ro & Ru & Ja & Avg. & Speed-up \\
    \midrule
    \multirow{1}*{MT} 
    & Transformer (Ours)  & 30.8 & 35.6 & 43.3 & 31.6 & 35.8 & 37.9 & 30.1 & 20.0  & 16.5 & 33.1 & - \\
    \midrule
    \multirow{14}*{AR} 
    & Transformer \cite{Inaguma_ICASSP2021} & 23.1 & - & 33.8 & - & - & - & - & - & - & - & - \\
    & \quad  + Seq-KD   & 24.4 & - & 34.6 & - & - & - & - & - & - & - & -    \\
    & Transformer  \cite{Inaguma_CORR2021} & 22.8 & 27.8 & 33.3 &  23.3 &  27.3 & - & - & - & - & - & -           \\
    & \quad  + Seq-KD & 24.3 & 28.9 & 34.5 & 24.2 & 28.4 & - & - & - & - &  - & - \\
    & Conformer \cite{Inaguma_CORR2021} & 25.0 & 30.5 & 35.5 &  25.4 &  29.7 & - & - & - & - & - & - \\
    & \quad + Seq-KD & 26.3 & 31.0 & 36.4 & 25.9 & 30.6 & - & - & - & - & -  & - \\
    \specialrule{0em}{1pt}{1pt}
    \cdashline{2-13}
    \specialrule{0em}{1pt}{1pt}
    & Fairseq ST \cite{Wang_AACL2020} & 22.7 & 27.2 & 32.9 & 22.7 & 27.3 & 28.1 & 21.9 & 15.3 & - & 24.8 & - \\
    & NeurST \cite{Zhao_ACL2021} & 22.8 & 27.4 & 33.3 & 22.9 & 27.2 & 28.7 & 22.2 & 15.1 & - & 24.9 & - \\
    & XSTNet \cite{Ye_interspeech2021} & 25.5 & 29.6 & 36.0 & 25.5 & 30.0 & 31.3 & 25.1 & 16.9 & - & 27.5 & - \\
    & STEMM \cite{Fang_ACL2022} & 25.6 & 30.3 & 36.1 & 25.6 & 30.1 &  31.0 & 24.3 & 17.1 & - & 27.5 & - \\
    & ConST \cite{Ye_NAACL2022} & 25.7 & 30.4 & 36.8 & 26.3 & 30.6 & 32.0 & 24.8 & 17.3 & - & 28.0 & - \\
    & M$^3$ST \cite{Cheng_Corr2022} & 26.4 & 31.0 & 37.2 & 26.6 & 30.9 & 32.8 & 25.4 & 18.3 & - & 28.6 & - \\
    \specialrule{0em}{1pt}{1pt}
    \cdashline{2-13}
    \specialrule{0em}{1pt}{1pt}
    & CTC-Aug ST (Ours) & 26.9 & 31.5 & 38.1 & 27.4 & 31.9 & 33.4 & 25.8 & \textbf{18.7} & 16.1 & 29.2 &  1.0$\times$ \\
    & \quad  + Seq-KD & \textbf{27.7} & \textbf{31.6} & \textbf{39.5} & \textbf{27.5} & \textbf{32.3} & \textbf{33.7} & \textbf{26.6} & \textbf{18.7} & \textbf{16.4} & \textbf{29.7} & 1.0$\times$ \\
    \midrule
    \multirow{6}*{NAR} 
    & CTC \cite{Inaguma_ICASSP2021}                   & 19.4 & - & 27.4 & - & - & - & - & - & - &	- & 20.84$\times$    \\
    & Orthros \cite{Inaguma_ICASSP2021}                 & 23.9 & - & 33.1 & - & - & - & - & - & - & -	& 2.39$\times$  \\
    \specialrule{0em}{1pt}{1pt}
    \cdashline{2-13}
    \specialrule{0em}{1pt}{1pt}
    & CTC \cite{Inaguma_CORR2021}  & 24.1 & 29.0 & 34.6 & 24.3 & 28.5 & - & - & - & - &  - & 13.83$\times$   \\
    & Orthros - CTC \cite{Inaguma_CORR2021}    & 25.3 & 30.4 & 36.2 &  25.4 &  29.9 & - & - & - & - &  - & 1.14$\times$ \\
    & Orthros - CMLM \cite{Inaguma_CORR2021} & 24.1 &  29.2 & 35.1 &  24.4 & 28.6 & - & - & - & - & - & 2.73$\times$ \\
    \specialrule{0em}{1pt}{1pt}
    \cdashline{2-13}
    \specialrule{0em}{1pt}{1pt}
    & CTC-NAST (Ours) & \textbf{27.3} & \textbf{31.8} & \textbf{38.9} & \textbf{27.7} & \textbf{32.3} & \textbf{33.3} & \textbf{26.1} & \textbf{18.9} & \textbf{16.2} & \textbf{29.5} & 5.67$\times$ \\
    \bottomrule
  \end{tabular}
  }
  \caption{BLEU scores on MuST-C corpora. The speed-up is calculated on the En-De corpus.}
  \label{mustc_bleu}
\end{table*}

However, this strategy results in the inconsistency between training and decoding, where the ground truth is unavailable during decoding.
To address this issue, \newcite{Qian_ACL2021} adaptively determine the replacement ratio depending on the current prediction accuracy. 
But it does not work for CTC-NAST, as shown in Appendix \ref{ablation_clm}.

Considering the long input sequence in ST, a lower ratio may not provide sufficient prompt, but a higher ratio may result in a severe gap between training and decoding.
Therefore, we limit that only the positions where a wrong prediction ($\arg \max p_t \ne \hat{\pi}_t$) occurs are replaced.
In this way, we enable the large ratio throughout the whole training process.
As the accuracy increases, more and more positions rely on the model's predictions, and the guidance to the fewer positions with errors always remains stable for better convergence.
We call this method curriculum learning mixing (CLM).

Finally, we smooth the ground truth to obtain a distribution similar to the CTC prediction, where the dominant probability is concentrated on the ground truth position, and the rest is evenly distributed among other tokens.

\subsection{Inference}

CTC-NAST is a fully parallel decoding model. 
The inference resembles the training process, except the CLM method is not used.
We employ greedy decoding, where CTC picks the tokens with maximum probability in each time-step, then removes the blanks and repeated tokens for final translation.

\section{Extension on AR model}

Now a natural question arises: can our method proposed for the NAR model be used to improve the AR model?
Our method produces better encoder representations for CTC prediction, but there is no evidence to demonstrate that the optimization of the CTC and the cross-entropy in the decoder are completely consistent.
Excessive optimization of the encoder may interfere with the learning of the decoder.

To answer it, we adopt these techniques to the encoder-decoder counterpart (called CTC-Aug ST), to investigate the effects of different architectures.
And the training loss is formulated as:
\begin{eqnarray}
    \mathcal{L} & = & \mathcal{L}_{\rm S2S} + \alpha_{\rm A} \cdot \mathcal{L}_{\rm CTC} + \alpha_{\rm T} \cdot \mathcal{L}_{\rm XCTC} \nonumber \\
    & + & \beta_{\rm A} \cdot \frac{1}{M}\sum_{m=1}^{m} \mathcal{L}_{\rm InterCTC}^m \nonumber \\
    & + & \beta_{\rm T} \cdot \frac{1}{N}\sum_{n=1}^{N} \mathcal{L}_{\rm InterXCTC}^n
\end{eqnarray}
where $\mathcal{L}_{\rm S2S}$ is the cross-entropy loss of the decoder.

\section{Experiments}

We evaluate our method on the MuST-C and Fisher-Callhome benchmarks. Details about the datasets and model settings are described in Appendix \ref{app_A}.

\begin{table*}[t!]
  \centering
  \resizebox{\textwidth}{!}{
  \normalsize
  \begin{tabular}{clcccccc}
    \toprule
    & \multirow{2}*{Model} & \multicolumn{3}{c}{Fisher} & \multicolumn{2}{c}{Callhome}  & \multirow{2}*{Speed-up} \\   
    \cmidrule(lr){3-5} \cmidrule(lr){6-7}
    & & dev & dev2 & test & devtest & evltest \\
    \midrule
    \multirow{1}*{MT} 
    & Transformer (Ours) & 64.50 & 65.20 & 63.35 & 32.21 & 31.58 & - \\
    \midrule
    \multirow{6}*{AR} 
    & Transformer + Seq-KD  \cite{Inaguma_ICASSP2021}           & - & - & 50.32 & - & 19.81 & -    \\
    & Transformer + Seq-KD  \cite{Inaguma_CORR2021}             & 51.10 & 51.40 & 50.80 & 19.60 & 19.20 & -  \\
    & Conformer + Seq-KD  \cite{Inaguma_CORR2021}             & 54.70 & 55.40 & 54.10 & 21.50 & 21.00 & -  \\
    & Transformer + MTL + ASR init. \cite{Chuang_ACL2021}             & 48.27 & 49.17 & 48.40 & 17.26 & 17.45 & -  \\
    \specialrule{0em}{1pt}{1pt}
    \cdashline{2-8}
    \specialrule{0em}{1pt}{1pt}
    & CTC-Aug ST (Ours) & 53.61 & 54.07 & 53.69 & 22.16 & 21.33 & 1.0$\times$  \\
    & \quad  + Seq-KD & \textbf{55.39} & \textbf{55.88} & \textbf{55.09} & \textbf{23.09} & \textbf{22.92} & 1.0$\times$  \\
    \midrule
    \multirow{10}*{NAR} 
    & CTC \cite{Inaguma_ICASSP2021}  & - & - & 45.97 & - & 15.91 & 20.84$\times$     \\
    \specialrule{0em}{1pt}{1pt}
    \cdashline{2-8}
    \specialrule{0em}{1pt}{1pt}
    & Conformer - CTC \cite{Inaguma_CORR2021}  & 51.00 & 51.60 & 50.80 & 18.00 & 18.70 & 11.80$\times$   \\
    & Orthros - CTC \cite{Inaguma_CORR2021}    & 54.00 & 54.80 & 54.10 & 21.00 & 20.80 & 1.09$\times$ \\
    & Orthros - CMLM \cite{Inaguma_CORR2021} & 51.30 & 52.20 & 51.20 & 20.90 & 20.40 & 2.70$\times$ \\
    \specialrule{0em}{1pt}{1pt}
    \cdashline{2-8}
    \specialrule{0em}{1pt}{1pt}
    & Transformer - CTC \cite{Chuang_ACL2021}  & 42.61 & 43.91 & 43.50 & 13.02 & 13.52 & 28.9$\times$  \\
    & CTC + MTL \cite{Chuang_ACL2021} & 44.45 & 45.23 & 44.92	& 14.20	& 14.19	& 28.9$\times$ \\
    \specialrule{0em}{1pt}{1pt}
    \cdashline{2-8}
    \specialrule{0em}{1pt}{1pt}
    & Mask - CTC \cite{Higuchi_ASRU2021} & 51.10 & 51.70 & 50.60 & 17.90 & 18.30 & - \\
    & Intermediate CTC \cite{Higuchi_ASRU2021} & 51.30 & 51.40 & 51.00 & 19.00 & 19.00 & - \\
    & Self-conditioned CTC \cite{Higuchi_ASRU2021} & 50.70 & 51.20 & 50.50 & 19.10 & 19.20 & - \\
    \specialrule{0em}{1pt}{1pt}
    \cdashline{2-8}
    \specialrule{0em}{1pt}{1pt}
    & CTC-NAST (Ours) & \textbf{55.21} & \textbf{55.92} & \textbf{54.71} & \textbf{23.43} & \textbf{23.30} & 4.10$\times$ \\
    \bottomrule
  \end{tabular}
  }
  \caption{BLEU scores on Fisher-Callhome corpus.}
  \label{fisher_bleu}
\end{table*}

\subsection{Main Results}

The results on the MuST-C corpora in Table \ref{mustc_bleu} show that our method significantly outperforms previous AR and NAR models.
We achieve remarkable gains for all language pairs.
Here we highlight several major breakthroughs:
i) CTC-Aug ST is shown to be effective for the AR models, which gains an average of 0.6 BLEU points over the previous best work even without the augmentation of sequence-level knowledge distillation (Seq-KD) data. 
Note that not all proposed methods are used in CTC-Aug ST (see Section \ref{effect_method}).
ii) Our CTC-NAST models achieve comparable or better performance to the powerful AR counterparts on all 9 language pairs, with a high speed-up of 5.67$\times$. Note that CTC-NAST achieves a higher speed-up under large batch sizes (see Section \ref{speedup}).
iii) Referring to Appendix \ref{reordering}, the En-Ja translation has a strong demand for reordering capability. Our method also works well on this challenging distant language pair, demonstrating the potential of CTC-NAST. 

Similar results on Fisher-Callhome are shown in Table \ref{fisher_bleu}.
Interestingly, the NAST model outperforms the AR counterpart with 0.3 $\sim$ 0.4 BLEU points on the out-of-domain Callhome sets.
We find that the AR models miss some segments when translating the long sentences, while the CTC-NAST models still guarantee good performance, as shown in Appendix \ref{oom}.
It demonstrates the robustness of our CTC-NAST model.

\subsection{Analysis}

Next, we study several interesting problems on MuST-C En-De and En-Ja datasets to investigate the effects on similar and distant languages. We present further analyses in Appendix \ref{app_ana}.

\begin{figure}[t!]
    \centering
    \begin{tikzpicture}
    \footnotesize{
    \begin{axis}[
    ymajorgrids,
    xmajorgrids,
    grid style=dashed,
    width=.5\textwidth,
    height=0.3\textwidth,
    legend columns=4,
    legend entries={ST, CTC-Aug ST, NAST, CTC-NAST},
    legend style={
      draw=none,
      line width=1pt,
    },
    legend style={at={(0.5,1.0)}, anchor=south,legend cell align=left,
    nodes={scale=0.8, transform shape}},
    xlabel=\footnotesize{Output Length},
    ylabel=\footnotesize{BLEU},  
    xmin=0,xmax=80,
    ymin=18,ymax=31,
    xtick={0,20,40,60,80},
    ytick={18,22,26,30},
    ylabel style={yshift=0.0em},
    xlabel style={yshift=-0.5em},
    yticklabel style={/pgf/number format/precision=0,/pgf/number format/fixed zerofill},
    scaled ticks=false,
    ]

    \addplot[myblue!80, line width=0.8pt] 
    file {table/length_ast_base.txt};
    
    \addplot[dred!80, line width=0.8pt] 
    file {table/length_ast_final.txt};
    
    \addplot[myblue!80, line width=0.8pt,dashed] 
    file {table/length_nast_base.txt};

    \addplot[dred!80, line width=0.8pt,dashed] 
    file {table/length_nast_final.txt};
    
    \end{axis}
    }
    \end{tikzpicture}
    \caption{BLEU scores over various output lengths.}
    \label{fig:length_bleu}
\end{figure}


\begin{table*}[t!]
  \centering
  \resizebox{\textwidth}{!}{
  \large
  \begin{tabular}{lcccccccccccc}
    \toprule
    \multirow{3}*{Model} & \multicolumn{4}{c}{En-De} & \multicolumn{4}{c}{En-Ja} & \multicolumn{3}{c}{Inference} & \multirow{3}*{Params.} \\  
    \cmidrule(lr){2-5} \cmidrule(lr){6-9} \cmidrule(lr){10-12}
    & \multicolumn{2}{c}{Raw} & \multicolumn{2}{c}{Seq-KD} & \multicolumn{2}{c}{Raw} & \multicolumn{2}{c}{Seq-KD} & \multirow{2}*{AR Times} & \multirow{2}*{NAR Times} & \multirow{2}*{Speed-up} & \\
    \cmidrule(lr){2-5} \cmidrule(lr){6-9} 
    & AR & NAR & AR & NAR & AR & NAR & AR & NAR & & & & \\
    \midrule 
    Base & 26.1 & - & 27.1 & - & 15.9 & - & 16.1 & - & 547.2 & - & - & $\sim$ 130M   \\
    \quad  + XCTC & 26.7 & 17.3 & 27.0 & 24.3 & 16.3 & 7.3 & 16.3 & 13.7 & 555.0 & 79.9 & 6.95$\times$ & $\sim$ 130M\\
    \quad\quad  + PAE & 26.9 & 19.6 & \textbf{27.7} & 25.7 & 16.1 & 8.5 & 16.4 & 14.9  & 545.0 & 84.1 & 6.48$\times$ & $\sim$ 140M \\
    \quad\quad\quad  + CLA & 26.8 & 19.1 & 27.3 & 26.2 & \textbf{16.6} & 10.0 & 16.4 & 15.3 & 565.6 & 91.8 & 6.16$\times$ & $\sim$ 150M \\
    \quad\quad\quad  + CLM & 26.6 & 25.7 & 27.5 & \textbf{27.4} & 14.4 & 14.3 & \textbf{16.6} & 16.1  & 543.1 & 82.3 & 6.60$\times$ & $\sim$ 140M \\
    \quad\quad\quad  + CLA + CLM & \textbf{27.0} & \textbf{25.8} & 27.6 & 27.3 & 13.6 & \textbf{14.5} & 16.2 & \textbf{16.2}  & 575.0 & 96.2 & 5.98$\times$ & $\sim$ 150M\\
    \bottomrule
  \end{tabular}
  }
  \caption{The effects of our methods on AR and NAR models.}
  \label{AR_NAR_mustc_bleu}
\end{table*}

\subsubsection{Performance over Sentence Lengths}

Figure \ref{fig:length_bleu} shows the results of the AR and NAR models with and without the proposed methods on the MuST-C En-De corpus with respect to output lengths.
The base NAR model performs much worse than AR counterpart.
But interestingly, unlike the ST model, which has an outlier as sentence length increases, the NAST model maintains stable performance.
This is similar to the results on Fisher-Callhome in Appendix \ref{oom}.

Our methods bring remarkable gains over different lengths for both AR and NAR models, leading to comparable translation quality when the length is less than 60.
In particular, CTC-NAST performs even better than AR models when the length is less than 30.
However, the performance gap increases with sentence length.
We speculate that very long input acoustic features make it more difficult to model semantic information.
Future work \cite{xu2023pds} can focus on enhancing the ability to handle complex acoustic encoding.

\subsubsection{Effects of Each Method}
\label{effect_method}

We compare the results of each method on AR and NAR models in Table \ref{AR_NAR_mustc_bleu}.
More detailed ablation studies of CLA and CLM are presented in Appendix \ref{app_ab}.
The base AR model is trained with auxiliary loss, where CTC on top of the acoustic encoder learns to predict the source text.
Interestingly, there are different effects on different models, languages, and training data.
All methods are lightweight in both computational cost and parameter quantity.

Introducing the XCTC loss and PAE method achieves better performance in nearly all settings.
CLA does not work well on the similar En-De language pair due to the less reordering requirement, but stable improvements on the distant En-Ja language pair. 
The remarkable results of CLM demonstrate that an adaptive training strategy is important for better convergence of NAR models \cite{Qian_ACL2021}.

However, CLM leads to slightly worse or better results for AR models trained on Seq-KD data.
We conclude that the optimization of XCTC loss in the encoder interferes with the learning of cross-entropy loss in the decoder.
Although the XCTC achieves good performance, it does not contribute to the final inference in the encoder-decoder framework.
In addition, the performance of the AR model trained on raw En-Ja data drops terribly.
Raw data distribution is difficult to learn by CTC, especially for distant En-Ja language pair.
In this case, the CLM always provides ground truth in a high ratio to mix, leading to overfitting on the training set and worse performance during inference.
Therefore, we only use XCTC and PAE on AR models for stable improvements.

We also notice that the simplified data distribution is crucial for achieving optimal performance with the NAST model.
Specifically, the base NAR models, when trained on raw data, significantly underperform models trained on Seq-KD data, with a gap of about 7 BLEU points. 
By combining proposed methods, we develop a powerful NAR model that narrows the gap to within 2 BLEU points. 
This result highlights the robustness of CTC-NAST, even in the presence of complex data distributions.


\begin{table}[t!]
  \centering
  \resizebox{\columnwidth}{!}{
  \large
  \begin{tabular}{p{0.4mm}lcccccc}
    \toprule
    \multicolumn{2}{l}{\multirow{2}{*}[-3pt]{Model}}
    & \multicolumn{3}{c}{En-De} & \multicolumn{3}{c}{En-Ja} \\  
    \cmidrule(lr){3-5} \cmidrule(lr){6-8}
    & & sub & del & ins & sub & del & ins \\
    \midrule
    \multirow{2}*{\begin{rotatebox}{90}{AR}\end{rotatebox}} 
    & Base & 31.8 & 12.2 & \textbf{12.5} & 44.6 & \textbf{19.3} & 16.9 \\
    & \quad + XCTC-Aug & \textbf{31.4} & \textbf{12.0} & \textbf{12.5} & \textbf{43.9} & 19.6 & \textbf{15.9} \\
    \specialrule{0em}{1pt}{1pt}
    \cdashline{1-8}
    \specialrule{0em}{1pt}{1pt}
    \multirow{5}*{\begin{rotatebox}{90}{NAR}\end{rotatebox}}
    & Base & 32.0 & 14.4 & \textbf{10.7} & 42.8 & 22.8 & \textbf{12.8} \\
    & \quad  + PAE & 31.6 & 13.2 & 11.4 & 43.2 & 21.1 & 14.4 \\
    & \quad\quad  + CLA & 31.4 & 12.9 & 11.7 & 43.6 & \textbf{20.3} & 14.8 \\
    & \quad\quad  + CLM & \textbf{30.8} & \textbf{12.8} & 11.3 & \textbf{42.1} & 21.2 & 13.7 \\
    & \quad\quad  + CLA + CLM & 30.9 & \textbf{12.8} & 11.4 & \textbf{42.1} & 21.2 & 14.0 \\
    \bottomrule
  \end{tabular}
  }
  \caption{Error analysis based on WERs that are split into substitution (sub), deletion (del), and insertion (ins) error rates.}
  \label{error_types}
\end{table}
\renewcommand{\thefootnote}{\arabic{footnote}}
\subsubsection{Error Analysis}

To identify the weakness of NAR generation, we measure the word error rates (WERs) of AR and NAR models on the MuST-C En-De and En-Ja datasets\footnote{Although WER is the metric for ASR, it helps to understand the error types of the translation results.}.
For a token in the target text, the sub error indicates that it is incorrectly translated, and the del error indicates that it is omitted.
The ins error indicates that the token not in the target text is translated.

High del error rates show that the dominant disadvantage of the NAST model is missing translation.
PAE relaxes the conditional independence assumption, giving better results for En-De but increased sub errors for En-Ja.
We speculate that this is because poor CTC prediction introduces excessive errors.
CLA is particularly effective at reducing del errors, which is consistent with our motivation to relax the monotonic assumption.
And CLM reduces error propagation and improves the robustness of PAE, achieving consistent improvements.

However, the combination of our methods does not lead to a further reduction in del errors.
A possible reason is that the inconsistent learning between CLA and CLM limits the effect of the combination.
We will explore better methods to alleviate the missing translation in the future.



\begin{figure}[t!]
    \centering
    \begin{tikzpicture}
    \footnotesize{
    \begin{axis}[
    ymajorgrids,
    xmajorgrids,
    grid style=dashed,
    width=.5\textwidth,
    height=0.35\textwidth,
    legend columns=2,
    legend entries={Beam-1,Beam-5},
    legend style={
      draw=none,
      line width=1pt,
    },
    legend style={at={(0.5,1.0)}, anchor=south,
    nodes={scale=0.8, transform shape}},
    xmode=normal,
    xlabel=\footnotesize{Sentences per Batch},
    ylabel=\footnotesize{Speed-up},  
    ymin=4.5,ymax=8.5,
    xtick distance=1,
    enlarge x limits=0.05,
    symbolic x coords={1,4,8,16,32},
    ytick={4,5,6,7,8},
    ylabel style={yshift=0.0em},
    xlabel style={yshift=0.0em},
    scaled ticks=false,
    ]
    
    \addplot+[ugreen!80, mark=square, line width=1pt] file {table/sp_ar1.txt};
    
    \addplot+[dred!80, mark=triangle, line width=1pt] file {table/sp_ar5.txt};
    
    \end{axis}
    }
    \end{tikzpicture}
    \caption{Speed-up under different settings.}
    \label{fig:speed_up}
\end{figure}

\subsubsection{Speed-up vs. Batch Size}
\label{speedup}
We examine the speed-up compared to AR models under different batch sizes and beam sizes in Figure \ref{fig:speed_up}.
Our CTC-NAST model consistently maintains a high speed-up, even with a large batch size of 32.
The performance of NAR and AR models is comparable when using a beam size of 1, while our NAR model is more than $5\times$ faster.
In addition, our encoder-only design simplifies the inference process, eliminating the need for length prediction or iterative refinement.
One promising direction is to develop effective encoding methods that can bridge the length gap between acoustic features and text.
This has the potential to reduce the computational cost caused by long sequence modeling.

\section{Conclusion}

Aiming to combine E2E ST and NAR generation, we propose CTC-NAST, which consists of only two CTC-guided encoders for source and target text prediction, respectively.
We identify and address several challenges of CTC-NAST: conditional independence assumption, monotonic assumption, and poor convergence.
In this way, our CTC-NAST model outperforms the previous best AR models by 0.9 BLEU points.
We believe that we are the first to present a NAST model that achieves comparable or better performance than strong AR counterparts.

\section*{Limitations}

Although our CTC-NAST model achieves excellent performance, there are still some underlying challenges that remain in the follow-up of our work.
Here are some limitations that we intend to resolve in the future:
\begin{itemize}
    \item The better designs of reordering augmentation and training strategy. Although the proposed CLA and CLM approaches achieve good results by alleviating the monotonic assumption and relieving the modeling burden, combing them can not bring remarkable improvement. More importantly, these two methods fail to stable improvements in encode-decoder architecture. This drives us to investigate the interference of the optimizations between CTC and cross-entropy.
    \item Combination with the pre-training or multitask learning. 
    Although our methods bring remarkable gains on both AR and NAR models, we do not explore the utilization of external data resources.
    Although we can use the pre-trained models directly, we expect more effective methods in future work. 
    Theoretically, we need to design NAR ASR and MT models that share the same or similar architectures with the acoustic encoder and textual encoder, respectively. 
    In this way, the NAST model bridges the gap between pre-training and fine-tuning and has more potential for better performance.
    \item The potential risk for unwritten languages. 
    In our work, we assume that transcription is always available, which is consistent with almost previous studies. 
    Although some datasets have no transcription, we can use a well-trained ASR model to generate pseudo labels. 
    However, it is hard to handle speech translation from unwritten source speech. The supervision of source text is very important for our model. Therefore, we need to develop better methods for stable training.
\end{itemize}

\section*{Acknowledgement}

The authors would like to thank anonymous reviewers for their insightful comments.
This work was supported in part by the National Science Foundation of China (No. 62276056), the National Key R\&D Program of China, the China HTRD Center Project (No. 2020AAA0107904), the Natural Science Foundation of Liaoning Province of China (2022-KF-16-01), the Yunnan Provincial Major Science and Technology Special Plan Projects (No. 202103AA080015), the Fundamental Research Funds for the Central Universities (Nos. N2216016, N2216001, and N2216002), and the Program of Introducing Talents of Discipline to Universities, Plan 111 (No. B16009).

\bibliography{anthology,custom}
\bibliographystyle{acl_natbib}

\appendix
\label{sec:appendix}

\section{Experimental Settings}
\label{app_A}
\subsection{Datasets and Preprocessing}

We conduct experiments on the MuST-C \cite{GangiCBNT19} and Fisher-Callhome ST \cite{Post_IWSLT2013} datasets.
MuST-C is a multilingual speech translation corpus extracted from TED lectures.
We test our method on all MuST-C v1 corpora: English (En) to German (De), Spanish (Es), French (Fr), Italian (It), Dutch (Nl), Portuguese (Pt), Romanian (Ro) and Russian (Ru).
In addition, we also investigate the results of the distant language pair English-Japanese (En-Ja) corpus in the MuST-C v2 dataset.
We select (and tune) the model on the dev set (Dev) and report the results on the tst-COMMON set (Test).

Fisher-Callhome is a Spanish-English speech-to-text translation dataset with 138k text pairs.
This corpus contains 170 hours of Spanish conversational telephone speech, as well as Spanish transcripts and English translations.
Following the recipe of ESPnet \cite{Inaguma_ACL2020}, we lowercase all texts, and remove all punctuation marks except apostrophes.
We select (and tune) the model on the Fisher-dev set, and report the results on the Fisher-\{dev, dev2, test\} and Callhome-\{devtest, evltest\} sets.

Following the preprocessing recipes in the fairseq toolkit\footnote{\href{https://github.com/pytorch/fairseq}{https://github.com/pytorch/fairseq}}, we remove utterances with more than 3,000 frames or less than 5 frames.
We extract the 80-channel Mel filter bank features by a window size of 25ms with a stride of 10ms.
The text is tokenized using the scripts of Moses \cite{Koehn_ACL2007} except that the Japanese text uses MeCab\footnote{\href{https://github.com/taku910/mecab}{https://github.com/taku910/mecab}}.
We learn SentencePiece\footnote{\href{https://github.com/google/sentencepiece}{https://github.com/google/sentencepiece}} segmentation with a size of 10,000 for MuST-C datasets.
We use a shared vocabulary for the source and target languages for MuST-C v1 corpora, the independent vocabulary for the En-Ja corpus.
And we use a shared vocabulary with a size of 1, 000 for Fisher-Callhome datasets.

\subsection{Model Settings}

We implement our method based on the fairseq toolkit \cite{Ott_NAACL2019}.
We use the Adam optimizer with $\beta_1=0.9, \beta_2=0.98$, and adopt the default learning schedule in fairseq.
We apply dropout with a rate of 0.15 and label smoothing of 0.1 for regularization.

Following previous studies on NAR models, our model is trained by sequence-level knowledge distillation (Seq-KD) \cite{Kim_EMNLP2016} data generated by a small MT model with a beam size of 5.
Our NAST model consists of an acoustic encoder with 12 Conformer layers and a textual encoder with 12 Transformer layers.
Each layer comprises 512 hidden units, 8 attention heads, and 2048 feed-forward sizes.
We use PAE in layers 6 and 9 in both the acoustic encoder and the textual encoder.
In multitask learning, the weights of $\alpha_A$, $\alpha_T$, $\beta_A$ and $\beta_T$ are all set to 1.
We start the cross-layer attention from layer 4 in the textual encoder and take the representation output from layer 3 as the key and value.
The ratio for curriculum learning mixing is set to 0.8.

We extend our method to the encoder-decoder model with similar settings, where the textual encoder has 6 Transformer layers and the decoder has 6 layers.
In this way, we control the model parameters to about 150M for fair comparisons.
The weights of $\alpha_A$ and $\alpha_T$ are set to 0.2, and the weights of $\beta_A$ and $\beta_T$ are to 0.1.
We use PAE in layer 4 in the textual encoder.
We start the cross-layer attention from layer 3 and take the representation output from layer 2 as the key and value.

During inference, we average the model parameters on the best 10 checkpoints based on the performance of the development set.
We use beam search with a beam size of 5 for the AR model.
The decoding speed is measured on the test set with a batch size of 1 on an Nvidia A100 80GB GPU.
We run 5 times to calculate the average time.
We report case-sensitive SacreBLEU \cite{Post_wmt18} on the MuST-C datasets and case-insensitive SacreBLEU on the Fisher-Callhome dataset for standardization comparison across papers.

\begin{table}[t!]
  \centering
  \small
  \begin{tabular}{lrr}
    \toprule
    \multirow{1}*{En-xx}
    & Raw & Seq-KD \\
    \midrule 
    De & 7.23 & 5.10 \\
    Es & 4.42 & 2.72 \\
    Fr & 5.51 & 2.80 \\ 
    It & 5.79 & 2.94 \\
    Nl & 6.18 & 4.16 \\ 
    Pt & 5.56 & 3.26 \\
    Ro & 5.22 & 2.95 \\ 
    Ru & 6.99 & 2.94 \\ 
    Ja & 14.01 & 15.21 \\
    \bottomrule
  \end{tabular}
  \caption{Reordering difficulty of MuST-C datasets.}
  \label{reordering_difficulty}
\end{table}

\section{More Analysis}
\label{app_ana}

\subsection{Reordering Difficulty}
\label{reordering}

Following the metric in \citet{Chuang_ACL2021}, we measure the reordering difficulties $\mathrm{R}_\pi$ on 9 language pairs of MuST-C datasets in Table \ref{reordering_difficulty}.
The higher the value of $\mathrm{R}_\pi$, the higher the reordering difficulty between texts from two languages, indicating the high demand for improved reordering capability.
The Seq-KD technique reduces the reordering difficulty by simplifying the data distribution, except for En-Ja.
The reason is that noisy data leads to poor MT performance on the En-Ja dataset.
On this distant language pair, our CTC-NAST model still achieves a high BLEU score of 16.2, which is comparable to the AR model with a small gap of only 0.2 BLEU points.

\begin{figure}[t!]
    \centering
    \begin{tikzpicture}
    \footnotesize{
    \begin{axis}[
    ymajorgrids,
    xmajorgrids,
    grid style=dashed,
    width=.5\textwidth,
    height=0.295\textwidth,
    legend columns=2,
    legend entries={ST Dev, NAST Dev, ST Evl, NAST Evl},
    legend style={
      draw=none,
      line width=1pt
    },
    legend style={at={(0.5,1.0)}, anchor=south, legend cell align=left,
    nodes={scale=0.8, transform shape}},
    xlabel=\footnotesize{Output Length},
    ylabel=\footnotesize{BLEU},  
    xmin=0,xmax=80,
    ymin=5,ymax=30,
    xtick={0,20,40,60,80},
    ytick={5,10,15,20,25,30},
    ylabel style={yshift=0.0em},
    xlabel style={yshift=-0.5em},
    yticklabel style={/pgf/number format/precision=0,/pgf/number format/fixed zerofill},
    scaled ticks=false,
    ]
    
    \addplot[blgreen!80, line width=0.8pt] 
    file {table/length_ast_caldev.txt};

    \addplot[blgreen!80, line width=0.8pt,dashed] 
    file {table/length_nast_caldev.txt};
    
    \addplot[pink!80, line width=0.8pt] 
    file {table/length_ast_calevl.txt};
    
    \addplot[pink!80, line width=0.8pt,dashed] 
    file {table/length_nast_calevl.txt};
 
    \end{axis}
    }
    \end{tikzpicture}
    \caption{BLEU scores over various output lengths of Callhome sets.}
    \label{fig:callhome_length_bleu}
\end{figure}

\subsection{Results on Out-of-domain Data}
\label{oom}

We also measure the BLEU scores of AR and NAR models under different output lengths on the Callhome sets in Figure \ref{fig:callhome_length_bleu}. 
Note that Callhome sets are out-of-domain because we only use the Fisher set for training.
Here, BLEU scores of the NAR model are better than those of the AR model in most cases of output length. 
In particular, when the output length is greater than 50, the performance of the AR model drops sharply, while the performance of the NAR model keeps stable. 
This demonstrates that our CTC-NAST has better robustness.

\subsection{Ablation Studies}
\label{app_ab}

To further verify the effectiveness of our proposed methods, we construct a series of ablation studies on MuST-C En-De and En-Ja datasets.

\begin{table}[t!]
  \renewcommand{\arraystretch}{0.7}
  \centering
  \resizebox{\columnwidth}{!}{
  \large
  \begin{tabular}{lcccc}

    \toprule
    \multirow{2}*{\footnotesize Model}
    & \multicolumn{2}{c}{\footnotesize En-De} & \multicolumn{2}{c}{\footnotesize En-Ja} \\  
    \cmidrule(lr){2-3} \cmidrule(lr){4-5}
    & \footnotesize dev & \footnotesize test & \footnotesize dev & \footnotesize test \\
    \midrule 
   \footnotesize Base   & \footnotesize 23.7 & \footnotesize 24.3 & \footnotesize 10.5 & \footnotesize 13.7     \\
    \quad + \footnotesize PAE  & \footnotesize 24.8 & \footnotesize 25.7 & \footnotesize 12.4 & \footnotesize 14.9   \\
    \quad\quad  + \footnotesize CLA & \footnotesize 25.1 & \footnotesize 25.8 & \footnotesize 12.1 & \footnotesize 15.3 \\
    \quad\quad\quad  \footnotesize {+ drop 0.1} & \footnotesize \textbf{25.4} & \footnotesize \textbf{26.2} & \footnotesize \textbf{12.7} & \footnotesize 15.3 \\
    \quad\quad\quad  \footnotesize {+ drop 0.2 \quad\quad} & \footnotesize 25.2 & \footnotesize 25.5 & \footnotesize 12.3 & \footnotesize {\textbf{15.6}} \\
    \bottomrule
  \end{tabular}
  }
  \caption{Ablation study of the CLA module.}
  \label{cla_mustc}
\end{table}

\begin{table}[t!]
  \centering
  \resizebox{\columnwidth}{!}{
  \large
  \begin{tabular}{lcccc}
    \toprule
    \multirow{2}*{Model}
    & \multicolumn{2}{c}{En-De} & \multicolumn{2}{c}{En-Ja} \\  
    \cmidrule(lr){2-3} \cmidrule(lr){4-5}
    & 0.5 & 0.8 & 0.5 & 0.8 \\
    \midrule 
    Base & 24.3 & 24.3 & 13.7 & 13.7    \\
    \quad + PAE & 25.7 & 25.7 & 14.9 & 14.9 \\
    \quad\quad  + Mixing & 26.7 & 26.6 & 15.6 & 15.7\\
    \quad\quad\quad  + Adaptive & 26.2 & 26.3 & 15.2 & 15.4\\
    \quad\quad\quad  + Only error & 26.7 & 27.1 & 15.8 & 15.9\\
    \quad\quad\quad  + Smooth & 26.7 & 26.6 & 15.8 & 15.6\\
    \quad\quad\quad + Only error + Smooth & \textbf{26.8} & \textbf{27.4} & \textbf{16.0} & \textbf{16.1} \\
    \bottomrule
  \end{tabular}
  }
  \caption{Ablation study of the CLM method under different mixing ratios and strategies.}
  \label{ratio_mustc}
\end{table}

\label{ablation_cla}
\noindent \textbf{Effects of CLA}
Table \ref{cla_mustc} shows the results of the CLA module.
CLA improves the reordering capability and complements the self-attention module.
However, using the CLA module naively brings only modest improvements.
We randomly drop the self-attention module with a probability of 0.1, which provides better regularization and robust improvements.
Note that the high drop probability may lead to insufficient training of the self-attention module.
These results demonstrate the effectiveness of the CLA module and drop-net technique.

\label{ablation_clm}
\noindent \textbf{Effects of CLM}
As shown in Table \ref{ratio_mustc}, the straightforward mixed training has produces remarkable gains with a ratio of 0.5 or 0.8 on both En-De and En-Ja datasets.
The adaptive strategy in NAR MT does not work in CTC-NAST.
This is because the sequence length of the input acoustic feature is very lengthy, and the decreased mixing ratio cannot provide enough cues to facilitate training.
For stable training, we only replace positions where wrong predictions arise.
In this manner, accurate positions solely rely on self-prediction, guaranteeing consistency between training and decoding.
Furthermore, we generate a smooth distribution akin to CTC prediction, in which the ground truth token has a high probability of 0.9, and the probabilities of other tokens sum to 0.1.
The combination of these two approaches results in additional and stable improvements.

We also calculate BLEU scores with various mixing ratios in Figure \ref{fig:clm_comparison}.
Our CLM approach is superior to the naive mixing method, particularly at a high ratio.
In this case, our approach incorporates more revisions solely for incorrect predictions, which facilitates the training process and guarantees consistency.

\begin{figure}[t!]
    \centering
    \begin{tikzpicture}   
    \footnotesize{
        \begin{groupplot}[
        ymajorgrids,
        xmajorgrids,
        grid style=dashed,
        width=.5\textwidth,
        legend columns=2,
        legend style={
        draw=none,
        line width=1pt,
        },
        legend style={at={(0.5,1.0)}, anchor=south, legend cell align=left,
        nodes={scale=0.8, transform shape}},
        xmode=normal,
        group style={
                group name=my fancy plots,
                group size=1 by 2,
                xticklabels at=edge bottom,
                vertical sep=0pt
            },
        xmin=0.45,xmax=0.85,
        xtick={0.5,0.6,0.7,0.8},
        xlabel style={yshift=0.0em},
        yticklabel style={/pgf/number format/precision=1,/pgf/number format/fixed zerofill},
        ]
        
        \nextgroupplot[ymin=26,ymax=27.7,
                       ytick={26.5,27,27.5},
                       axis x line*=top, 
                       ylabel style={at={(-0.15, 0.1)}}, 
                       ylabel=\footnotesize{BLEU},
                       axis y discontinuity=crunch,
                        width=.5\textwidth,
                       height=0.23\textwidth]
                       
        \addplot+[orange!80, mark=star , line width=1pt, dashed] plot coordinates {
        (0.5,26.67) 
        (0.6,26.96) 
        (0.7,26.57)
        (0.8,26.61)
        };
        \addlegendentry{En-De base}
        
        \addplot+[orange!80, mark=triangle*, mark options={solid,fill=orange!80}, line width=1pt] plot coordinates {
        (0.5,26.81) 
        (0.6,27.1) 
        (0.7,27.16)
        (0.8,27.39)
        };
        \addlegendentry{En-De ours}
        
        \addplot+[dblue!80, mark=*, mark options={solid,fill=dblue!80},line width=1pt,dashed] plot coordinates {
            (0.5,25.9)
        };
        \addlegendentry{En-Ja base}
        
        \addplot+[dblue!80, mark=diamond* , line width=1pt] plot coordinates {
            (0.5,25.9)
        };
        \addlegendentry{En-Ja ours} 
        
       \nextgroupplot[ymin=15.3,ymax=16.5,
                       ytick={15.5,16,16.5},
                       axis x line*=bottom,
                       xlabel=\footnotesize{Mixing ratio},
                        width=.5\textwidth,
                       height=0.2\textwidth]
                       
        \addplot+[dblue!80, mark=*, mark options={solid,fill=dblue!80},line width=1pt,dashed] plot coordinates {
        (0.5,15.59) 
        (0.6,15.71) 
        (0.7,15.7)
        (0.8,15.69)
        };
        
        \addplot+[dblue!80, mark=diamond* , mark options={fill=dblue!80},line width=1pt] plot coordinates {
        (0.5,15.95) 
        (0.6,15.78) 
        (0.7,16.19)
        (0.8,16.13)
        };             
        \end{groupplot}
        }
    \end{tikzpicture}
    \caption{BLEU scores of base mixing and our CLM method.}
    \label{fig:clm_comparison}
\end{figure}

\end{CJK*}
\end{document}

%% file: fig/fig1.tex
\definecolor{qblue}{RGB}{174,208,238}
\definecolor{qpurple}{RGB}{187,161,203}
\definecolor{qgreen}{RGB}{177,213,200}
\definecolor{qorange}{RGB}{243,166,148}
\definecolor{qpink}{RGB}{249,199,207}
\definecolor{qyellow}{RGB}{254,208,129}
\begin{figure*}
    \centering
\subfigure[]{
    \label{fig:a}
    \begin{minipage}[t]{0.22\linewidth}
    \centering
    \begin{tikzpicture}
    \tikzstyle{textonly} = [font=\scriptsize,align=center]
    \tikzstyle{sublayer} = [rectangle,draw,minimum width=2cm,rounded corners=3pt,align=center,inner sep=3pt,minimum height=0.5cm,font=\scriptsize]
    \tikzstyle{background} = [rectangle,rounded corners=3pt,minimum width=3.5cm,minimum height=3cm,font=\footnotesize,align=center];
    \tikzstyle{arrow} = [font=\Large,align=center];
    \node[textonly] (speech) at (0,0) {Speech Features};
    \node[background,draw=qblue,thick] (aeback) at ([yshift=1.68cm]speech.north) {} ;
    \node[sublayer,fill=qblue!70,draw=qblue,thick] (ae) at ([yshift=0.65cm]speech.north) {Acoustic Encoder};
    \node[sublayer,fill=qyellow!70,draw=qyellow,thick] (sctc) at ([yshift=0.65cm]ae.north) {Source CTC};
    \node[textonly] (srcsent) at ([yshift=0.95cm]sctc.north) {Source Sequence:\\ I say hello world};  
    \node[arrow](arw1) at ([yshift=0.36cm]sctc.north) {{ $\Updownarrow$ }};
    \node[textonly](srcctc) at ([xshift=0.5cm]arw1.east) {CTC Loss};
    \node[background,draw=qorange,thick] (teback) at ([yshift=1.75cm]srcsent.north) {} ;
    \node[sublayer,fill=qorange!70,draw=qorange,thick] (te) at ([yshift=0.7cm]srcsent.north) {Textual Encoder};
    \node[sublayer,fill=qgreen!70,draw=qgreen,thick] (tctc) at ([yshift=0.65cm]te.north) {Target CTC};    
    \node[textonly] (tgtsent) at ([yshift=0.95cm]tctc.north) {Target Sequence:\\
    Ich sage hallo welt}; 
    \node[arrow](arw2) at ([yshift=0.36cm]tctc.north) {{ $\Updownarrow$ }};
    \node[textonly](tgtctc) at ([xshift=0.5cm]arw2.east) {XCTC Loss};
    \draw[->,thick](speech.north)--(ae.south);
    \draw[->,thick](ae.north)--(sctc.south);
    \draw[->,thick](aeback.north)--(te.south);
    \draw[->,thick](te.north)--(tctc.south);  
\end{tikzpicture}
    \end{minipage}
    }
\subfigure[]{
\label{fig:b}
\begin{minipage}[t]{0.22\linewidth}
    \centering
    \begin{tikzpicture}
    \tikzstyle{textonly} = [font=\scriptsize,align=center];
    \tikzstyle{sublayer} = [rectangle,thick,minimum width=2cm,rounded corners=3pt,align=center,inner sep=3pt,minimum height=0.5cm,font=\scriptsize];
    \tikzstyle{background} = [rectangle,rounded corners=3pt,minimum width=3.3cm,minimum height=2.5cm,font=\footnotesize,align=center];
    \tikzstyle{arrow} = [font=\Large,align=center];    
    
    \node[textonly] (clainput) at (0,0) {};  

    \node[background,draw,thick,minimum height=2.3cm](kback) at ([yshift=1.3cm]clainput.north){};
    \node[sublayer,fill=qblue!70,draw=qblue] (sa1) at ([yshift=0.8cm]clainput.north) {Self-Attn};
    \node[textonly] (sa1k) at ([xshift=-0.25cm,yshift=-0.2cm]sa1.south) {};
    \node[textonly] (sa1v) at ([xshift=0.3cm]sa1k.east) {};
    \node[textonly] (sa1q) at ([xshift=0.3cm]sa1v.east) {};
    \node[sublayer,fill=qyellow!70,draw=qyellow] (ffn1) at ([yshift=0.7cm]sa1.north) {FFN};
    \node[textonly] (enlayk) at ([xshift=1cm,yshift=0.26cm]ffn1.north) {Layer \textsl{k}};
    \node[textonly] (lays) at ([yshift=0.8cm]ffn1.north) {\footnotesize ...};

    \node[background,draw,thick,minimum height=3.5cm](lback) at ([yshift=1.85cm]lays.north){};
    \node[background,draw,dotted,thick,fill=gray!10,minimum width=2.2cm,minimum height=0.8cm](lback) at ([yshift=0.7cm]lays.north){};
    
    \node[sublayer,fill=qblue!70,draw=qblue] (sa2) at ([yshift=0.8cm]lays.north) {Self-Attn};
    \node[textonly] (sa2k) at ([xshift=-0.25cm,yshift=-0.2cm]sa2.south) {};
    \node[textonly] (sa2v) at ([xshift=0.3cm]sa2k.east) {};
    \node[textonly] (sa2q) at ([xshift=0.3cm]sa2v.east) {};
    \node[sublayer,fill=qorange!70,draw=qorange] (cla) at ([yshift=0.8cm]sa2.north) {Cross-Layer Attn};
    \node[textonly] (clak) at ([xshift=-0.25cm,yshift=-0.2cm]cla.south) {};
    \node[textonly] (clav) at ([xshift=0.3cm]clak.east) {};
    \node[textonly] (claq) at ([xshift=0.3cm]clav.east) {};
    \node[sublayer,fill=qyellow!70,draw=qyellow] (ffn2) at ([yshift=0.7cm]cla.north) {FFN};
    \node[textonly] (enlayl) at ([xshift=1cm,yshift=0.32cm]ffn2.north) {Layer \textsl{j}};
    \node[textonly] (ffnout) at ([yshift=0.8cm]ffn2.north) {};
    \draw[->,thick](clainput.north)--(sa1.south);
    \draw[->,thick](sa1.north)--(ffn1.south);
    \draw[->,thick](ffn1.north)--(lays.south);
    \draw[->,thick](lays.north)--(sa2.south);
    \draw[->,thick](cla.north)--(ffn2.south);
    \draw[->,thick](ffn2.north)--(ffnout.south);
    \draw[->,thick,rounded corners=3pt]([yshift=0.24cm]clainput.north)--([yshift=0.24cm,xshift=1.2cm]clainput.north)--([xshift=1.2cm,yshift=0.15cm]sa1.north)--([yshift=0.15cm]sa1.north);
    \draw[->,thick]([yshift=0.32cm]clainput.north) .. controls +(east:0.5) and +(south:0.2) .. ([xshift=0.6cm]sa1.south);
    \draw[->,thick]([yshift=0.32cm]clainput.north) .. controls +(west:0.5) and +(south:0.2) .. ([xshift=-0.6cm]sa1.south);
    \draw[->,thick,rounded corners=3pt]([yshift=0.25cm]sa1.north)--([yshift=0.25cm,xshift=1.2cm]sa1.north)--([xshift=1.2cm,yshift=0.1cm]ffn1.north)--([yshift=0.1cm]ffn1.north);
    \draw[->,thick,rounded corners=3pt]([yshift=0.23cm]lays.north)--([yshift=0.23cm,xshift=1.2cm]lays.north)--([xshift=1.2cm,yshift=0.11cm]sa2.north)--([yshift=0.11cm]sa2.north);
    \draw[->,thick]([yshift=0.32cm]lays.north) .. controls +(east:0.5) and +(south:0.2) .. ([xshift=0.6cm]sa2.south);
    \draw[->,thick]([yshift=0.32cm]lays.north) .. controls +(west:0.5) and +(south:0.2) .. ([xshift=-0.6cm]sa2.south);
    \draw[-,thick,rounded corners=5pt,qorange] ([yshift=-0.16cm]lays.south) -- ([xshift=-1.4cm,yshift=-0.16cm]lays.south)-- ([xshift=-1.4cm,yshift=1.65cm]lays.south)--([xshift=-0.48cm,yshift=1.65cm]lays.south);
    \draw[->,thick,qorange]([xshift=-1cm,yshift=1.65cm]lays.south) .. controls +(east:0.5) and +(south:0.2) .. ([xshift=-0.5cm]cla.south);
    \draw[->,thick,rounded corners=3pt]([yshift=0.22cm]sa2.north)--([yshift=0.22cm,xshift=1.2cm]sa2.north)--([xshift=1.2cm,yshift=0.11cm]cla.north)--([yshift=0.11cm]cla.north);
    \draw[->,thick,qorange]([xshift=-0.5cm,yshift=1.65cm]lays.south) .. controls +(east:0.5) and +(south:0.2) .. (cla.south);
    \draw[-,thick,qorange](sa2.north)--([yshift=-0.24cm]cla.south);
    \draw[->,thick,qorange]([yshift=-0.25cm,xshift=-0.01cm]cla.south) .. controls +(east:0.5) and +(south:0.2).. ([xshift=0.6cm]cla.south);

    \draw[->,thick,rounded corners=3pt]([yshift=0.25cm]cla.north)--([yshift=0.25cm,xshift=1.2cm]cla.north)--([xshift=1.2cm,yshift=0.15cm]ffn2.north)--([yshift=0.15cm]ffn2.north);

\end{tikzpicture}
\end{minipage}
}
\subfigure[]{
\label{fig:c}
\begin{minipage}[t]{0.5\linewidth}
    \centering
        \begin{tikzpicture}
    \tikzstyle{textonly} = [font=\scriptsize,align=center];
    \tikzstyle{sublayer} = [rectangle,draw,minimum width=2.5cm,rounded corners=3pt,align=center,inner sep=3pt,minimum height=0.5cm,font=\scriptsize];
    \tikzstyle{embed} = [circle,minimum height=0.3cm,minimum width=0.3cm,fill=qpurple!70,inner sep=0pt];
    \tikzstyle{wordembd} = [rectangle,minimum width=0.5cm,align=center,inner sep=3pt,minimum height=2cm,font=\scriptsize]; 
    \tikzstyle{prob_rec} = [rectangle,draw,minimum width=0.55cm,minimum height=2cm,fill=white]; 
    \tikzstyle{probr} = [rectangle,fill=qblue!70,minimum height=0.3cm,minimum width=0.2cm,inner sep=0pt];
    \tikzstyle{probp} = [rectangle,fill=qorange!70,minimum height=0.3cm,minimum width=0.2cm,inner sep=0pt];
    \tikzstyle{background} = [rectangle,rounded corners=3pt,minimum width=4.2cm,minimum height=3.3cm];

    \node[background] (refback) at (0,0) {};
    \node[background,fill=gray!10,minimum height=2.6cm] (refback1) at ([yshift=1.3cm]refback.south) {};
    \node[background,fill=qblue!50,minimum height=2.4cm,minimum width=2.9cm,rounded corners=0pt] (prob1back) at ([xshift=0.55cm,yshift=-0.35cm]refback.center) {};

    \node[prob_rec](rclo1) at ([xshift=0.42cm]prob1back.west){};
    \node[prob_rec](rclo2) at ([xshift=0.4cm]rclo1.east){};
    \node[prob_rec](rclo3) at ([xshift=0.4cm]rclo2.east){};
    \node[prob_rec](rclo4) at ([xshift=0.4cm]rclo3.east){};
    
    \node[probr,minimum width=0.1cm,fill=white](probr11) at ([yshift=-0.3cm]rclo1.north){};
    \node[probr,minimum width=0.2cm,fill=white](probr12) at ([yshift=-0.2cm]probr11.south){};
    \node[probr,minimum width=0.15cm,fill=white](probr13) at ([yshift=-0.2cm]probr12.south){};
    \node[probr,minimum width=0.38cm](probr14) at ([yshift=-0.2cm]probr13.south){};
    \node[probr,minimum width=0.12cm,fill=white](probr15) at ([yshift=-0.2cm]probr14.south){};
    \node[textonly](prob1c1) at ([yshift=-0.2cm]probr13.south) {0.8};
    \node[probr,minimum width=0.1cm,minimum height=0.1cm](probr1m1) at (probr11.center){};
    \node[probr,minimum width=0.2cm,minimum height=0.2cm](probr1m2) at (probr12.center){};
    \node[probr,minimum width=0.15cm,minimum height=0.15cm](probr1m3) at (probr13.center){};
    \node[probr,minimum width=0.12cm,minimum height=0.12cm](probr1m4) at (probr15.center){};

    \node[probr,minimum width=0.2cm,fill=white](probr21) at ([yshift=-0.3cm]rclo2.north){};
    \node[probr,minimum width=0.38cm](probr22) at ([yshift=-0.2cm]probr21.south){};
    \node[probr,minimum width=0.2cm,fill=white](probr23) at ([yshift=-0.2cm]probr22.south){};
    \node[probr,minimum width=0.1cm,fill=white](probr24) at ([yshift=-0.2cm]probr23.south){};
    \node[probr,minimum width=0.12cm,fill=white](probr25) at ([yshift=-0.2cm]probr24.south){};
    \node[textonly](prob1c2) at ([yshift=-0.2cm]probr21.south) {0.7};
    \node[probr,minimum width=0.2cm,minimum height=0.2cm](probr2m1) at (probr21.center){};
    \node[probr,minimum width=0.2cm,minimum height=0.2cm](probr2m2) at (probr23.center){};
    \node[probr,minimum width=0.1cm,minimum height=0.1cm](probr2m3) at (probr24.center){};
    \node[probr,minimum width=0.15cm,minimum height=0.15cm](probr2m4) at (probr25.center){};
    
    \node[probr,minimum width=0.1cm,fill=white](probr31) at ([yshift=-0.3cm]rclo3.north){};
    \node[probr,minimum width=0.26cm,fill=white](probr32) at ([yshift=-0.2cm]probr31.south){};
    \node[probr,minimum width=0.16cm,fill=white](probr33) at ([yshift=-0.2cm]probr32.south){};
    \node[probr,minimum width=0.12cm,fill=white](probr34) at ([yshift=-0.2cm]probr33.south){};
    \node[probr,minimum width=0.38cm](probr35) at ([yshift=-0.2cm]probr34.south){};
    \node[textonly](prob1c3) at ([yshift=-0.2cm]probr34.south) {0.6};
    \node[probr,minimum width=0.1cm,minimum height=0.1cm](probr3m1) at (probr31.center){};
    \node[probr,minimum width=0.14cm,minimum height=0.14cm](probr3m2) at (probr32.center){};
    \node[probr,minimum width=0.2cm,minimum height=0.2cm](probr3m3) at (probr33.center){};
    \node[probr,minimum width=0.12cm,minimum height=0.12cm](probr3m4) at (probr34.center){};

    \node[probr,minimum width=0.16cm,fill=white](probr41) at ([yshift=-0.3cm]rclo4.north){};
    \node[probr,minimum width=0.25cm,fill=white](probr42) at ([yshift=-0.2cm]probr41.south){};
    \node[probr,minimum width=0.38cm](probr43) at ([yshift=-0.2cm]probr42.south){};
    \node[probr,minimum width=0.12cm,fill=white](probr44) at ([yshift=-0.2cm]probr43.south){};
    \node[probr,minimum width=0.2cm,fill=white](probr45) at ([yshift=-0.2cm]probr44.south){};
    \node[textonly](prob1c4) at ([yshift=-0.2cm]probr42.south) {0.7};
    \node[probr,minimum width=0.13cm,minimum height=0.13cm](probr3m1) at (probr41.center){};
    \node[probr,minimum width=0.22cm,minimum height=0.22cm](probr3m2) at (probr42.center){};
    \node[probr,minimum width=0.12cm,minimum height=0.12cm](probr3m3) at (probr44.center){};
    \node[probr,minimum width=0.16cm,minimum height=0.16cm](probr3m4) at (probr45.center){};

    \node[textonly](multi1) at ([xshift=-0.25cm]rclo1.west){$\times$};
    \node[wordembd](wodembd1) at ([xshift=-0.5cm]rclo1.west){};
    \node[embed](w11) at ([yshift=-0.3cm]wodembd1.north){};
    \node[embed](w12) at ([yshift=-0.2cm]w11.south){};
    \node[embed](w13) at ([yshift=-0.2cm]w12.south){};
    \node[embed](w14) at ([yshift=-0.2cm]w13.south){};
    \node[embed](w15) at ([yshift=-0.2cm]w14.south){}; 

    \node[textonly](wod11) at ([xshift=-0.3cm]w11.west){$\phi$};
    \node[textonly](wod12) at ([xshift=-0.3cm]w12.west){Ich};
    \node[textonly](wod13) at ([xshift=-0.3cm]w13.west){Sage};
    \node[textonly](wod14) at ([xshift=-0.3cm]w14.west){Hallo};
    \node[textonly](wod15) at ([xshift=-0.3cm]w15.west){Welt}; 

    \node[background,fill=qpink!50,minimum height=0.4cm] (ctcoutback) at ([yshift=-0.48cm]refback.north){};
    \node[embed,fill=qyellow,minimum height=0.35cm,minimum width=0.35cm,draw,thick](out1) at ([yshift=0.15cm,xshift=0.18cm]ctcoutback.west){\scriptsize \textbf {\textsl{1}} };
    \node[embed,fill=qblue,draw=qblue](ow11) at ([yshift=0.5cm]rclo1.north){};
    \node[embed,fill=qblue,draw=qblue](ow12) at ([yshift=0.5cm]rclo2.north){};
    \node[embed,fill=qblue,draw=qblue](ow13) at ([yshift=0.5cm]rclo3.north){};
    \node[embed,fill=qblue,draw=qblue](ow14) at ([yshift=0.5cm]rclo4.north){};

    \draw[->,thick](rclo1)--(ow11);
    \draw[->,thick](rclo2)--(ow12);
    \draw[->,thick](rclo3)--(ow13);
    \draw[->,thick](rclo4)--(ow14);

    \node[background,fill=qpink!50,minimum height=0.4cm] (dcloutback) at ([yshift=0.16cm]refback.north){};
  
    \node[embed,fill=qorange,draw=red,thick](dcl1) at ([yshift=0.5cm]ow11.north){};
    \node[embed,fill=qblue,draw=qblue,thick](dcl2) at ([yshift=0.5cm]ow12.north){};
    \node[embed,fill=qblue,draw=qblue,thick](dcl3) at ([yshift=0.5cm]ow13.north){};
    \node[embed,fill=qorange,draw=red,thick](dcl4) at ([yshift=0.5cm]ow14.north){};

    \node[background,fill=gray!10,minimum height=2.6cm] (paeback) at ([yshift=2.3cm]refback.north) {};
    \node[background,fill=qorange!50,minimum height=2.4cm,minimum width=2.9cm,rounded corners=0pt] (prob2back) at ([xshift=0.55cm]paeback.center) {};
    

    \node[prob_rec,draw=red,dashed,thick](pclo1) at ([xshift=0.42cm]prob2back.west){};
    \node[prob_rec](pclo2) at ([xshift=0.4cm]pclo1.east){};
    \node[prob_rec](pclo3) at ([xshift=0.4cm]pclo2.east){};
    \node[prob_rec,draw=red,dashed,thick](pclo4) at ([xshift=0.4cm]pclo3.east){};

    \node[probp,minimum width=0.38cm](probp11) at ([yshift=-0.3cm]pclo1.north){};
    \node[probp,minimum width=0.1cm,fill=white](probp12) at ([yshift=-0.2cm]probp11.south){};
    \node[probp,minimum width=0.1cm,fill=white](probp13) at ([yshift=-0.2cm]probp12.south){};
    \node[probp,minimum width=0.1cm,fill=white](probp14) at ([yshift=-0.2cm]probp13.south){};
    \node[probp,minimum width=0.1cm,fill=white](probp15) at ([yshift=-0.2cm]probp14.south){};
    \node[textonly](prob2c1) at ([yshift=-0.3cm]pclo1.north) {0.9};
    \node[probp,minimum width=0.15cm,minimum height=0.15cm](probp1m1) at (probp12.center){};
    \node[probp,minimum width=0.15cm,minimum height=0.15cm](probp1m2) at (probp13.center){};
    \node[probp,minimum width=0.15cm,minimum height=0.15cm](probp1m3) at (probp14.center){};
    \node[probp,minimum width=0.15cm,minimum height=0.15cm](probp1m4) at (probp15.center){};
    
    \node[probp,minimum width=0.1cm,fill=white](probp21) at ([yshift=-0.3cm]pclo2.north){};
    \node[probp,minimum width=0.38cm](probp22) at ([yshift=-0.2cm]probp21.south){};
    \node[probp,minimum width=0.1cm,fill=white](probp23) at ([yshift=-0.2cm]probp22.south){};
    \node[probp,minimum width=0.1cm,fill=white](probp24) at ([yshift=-0.2cm]probp23.south){};
    \node[probp,minimum width=0.1cm,fill=white](probp25) at ([yshift=-0.2cm]probp24.south){};
    \node[textonly](prob2c2) at ([yshift=-0.2cm]probp21.south) {0.9};
    \node[probp,minimum width=0.15cm,minimum height=0.15cm](probp2m1) at (probp21.center){};
    \node[probp,minimum width=0.15cm,minimum height=0.15cm](probp2m2) at (probp23.center){};
    \node[probp,minimum width=0.15cm,minimum height=0.15cm](probp2m3) at (probp24.center){};
    \node[probp,minimum width=0.15cm,minimum height=0.15cm](probp2m4) at (probp25.center){};
    
    \node[probp,minimum width=0.1cm,fill=white](probp31) at ([yshift=-0.3cm]pclo3.north){};
    \node[probp,minimum width=0.1cm,fill=white](probp32) at ([yshift=-0.2cm]probp31.south){};
    \node[probp,minimum width=0.1cm,fill=white](probp33) at ([yshift=-0.2cm]probp32.south){};
    \node[probp,minimum width=0.1cm,fill=white](probp34) at ([yshift=-0.2cm]probp33.south){};
    \node[probp,minimum width=0.38cm](probp35) at ([yshift=-0.2cm]probp34.south){};
    \node[textonly](prob2c3) at ([yshift=-0.2cm]probp34.south) {0.9};
    \node[probp,minimum width=0.15cm,minimum height=0.15cm](probp3m1) at (probp31.center){};
    \node[probp,minimum width=0.15cm,minimum height=0.15cm](probp3m2) at (probp32.center){};
    \node[probp,minimum width=0.15cm,minimum height=0.15cm](probp3m3) at (probp33.center){};
    \node[probp,minimum width=0.15cm,minimum height=0.15cm](probp3m4) at (probp34.center){};

    \node[probp,minimum width=0.1cm,fill=white](probp41) at ([yshift=-0.3cm]pclo4.north){};
    \node[probp,minimum width=0.1cm,fill=white](probp42) at ([yshift=-0.2cm]probp41.south){};
    \node[probp,minimum width=0.1cm,fill=white](probp43) at ([yshift=-0.2cm]probp42.south){};
    \node[probp,minimum width=0.38cm](probp44) at ([yshift=-0.2cm]probp43.south){};
    \node[probp,minimum width=0.1cm,fill=white](probp45) at ([yshift=-0.2cm]probp44.south){};
    \node[textonly](prob2c4) at ([yshift=-0.2cm]probp43.south) {0.9};
    \node[probp,minimum width=0.15cm,minimum height=0.15cm](probp4m1) at (probp41.center){};
    \node[probp,minimum width=0.15cm,minimum height=0.15cm](probp4m2) at (probp42.center){};
    \node[probp,minimum width=0.15cm,minimum height=0.15cm](probp4m3) at (probp43.center){};
    \node[probp,minimum width=0.15cm,minimum height=0.15cm](probp4m4) at (probp45.center){};

    \node[textonly](multi2) at ([xshift=-0.25cm]pclo1.west){$\times$};
    \node[wordembd](wodembd2) at ([xshift=-0.5cm]pclo1.west){};
    \node[embed](w21) at ([yshift=-0.3cm]wodembd2.north){};
    \node[embed](w22) at ([yshift=-0.2cm]w21.south){};
    \node[embed](w23) at ([yshift=-0.2cm]w22.south){};
    \node[embed](w24) at ([yshift=-0.2cm]w23.south){};
    \node[embed](w25) at ([yshift=-0.2cm]w24.south){}; 
    
    \node[textonly](wod21) at ([xshift=-0.3cm]w21.west){$\phi$};
    \node[textonly](wod22) at ([xshift=-0.3cm]w22.west){Ich};
    \node[textonly](wod23) at ([xshift=-0.3cm]w23.west){Sage};
    \node[textonly](wod24) at ([xshift=-0.3cm]w24.west){Hallo};
    \node[textonly](wod25) at ([xshift=-0.3cm]w25.west){Welt}; 

    \node[embed,fill=qorange,draw=red,thick](ow21) at ([yshift=-0.5cm]pclo1.south){};
    \node[embed,fill=qorange,draw=qorange](ow22) at ([yshift=-0.5cm]pclo2.south){};
    \node[embed,fill=qorange,draw=qorange](ow23) at ([yshift=-0.5cm]pclo3.south){};
    \node[embed,fill=qorange,draw=red,thick](ow24) at ([yshift=-0.5cm]pclo4.south){};

    \draw[->,thick,red](pclo1)--(ow21);
    \draw[->,thick](pclo2)--(ow22);
    \draw[->,thick](pclo3)--(ow23);
    \draw[->,thick,red](pclo4)--(ow24);
    \draw[-,very thick,dotted](ow13)--(dcl3);
    \draw[-,very thick,dotted](ow12)--(dcl2);
    \draw[-,draw=red,very thick,dotted](ow21)--(dcl1);
    \draw[-,draw=red,very thick,dotted](ow24)--(dcl4);

    \node[background,draw=gray,dotted,semithick,minimum height=3.66cm] (delback) at ([yshift=-0.51cm]paeback.center) {};
    \node[embed,fill=qyellow,minimum height=0.35cm,minimum width=0.35cm,draw,thick](out2) at ([yshift=0.15cm,xshift=0.18cm]dcloutback.west){\scriptsize \textbf {\textsl{2}}};  

\node[background,fill=qpink!50,minimum height=0.5cm, minimum width=2.5cm] (emback) at ([xshift=3.8cm,yshift=0.3cm]refback.south) {};
    \node[embed,fill=qyellow,draw=qyellow] (e2) at ([xshift=-0.3cm]emback.center) {};
    \node[embed,fill=qyellow,draw=qyellow] (e1) at ([xshift=-0.4cm]e2.west) {};
    \node[embed,fill=qyellow,draw=qyellow] (e3) at ([xshift=0.3cm]emback.center) {};
    \node[embed,fill=qyellow,draw=qyellow] (e4) at ([xshift=0.4cm]e3.east) {};
    \node[sublayer,fill=qgreen!70,draw=qgreen,thick] (ln) at ([yshift=0.8cm]emback.north) {Layer Norm};
    \node[sublayer,fill=qblue!70,draw=qblue,thick] (sm) at ([yshift=0.8cm]ln.north) {Softmax};
    \node[embed,fill=qyellow,minimum height=0.35cm,minimum width=0.35cm,draw,thick](out21) at ([yshift=0.45cm,xshift=-1.2cm]sm.north){\scriptsize \textbf {\textsl{1}} };
    \node[textonly] (or) at ([xshift=0.15cm]out21.east){\textbf{or}};
    \node[embed,fill=qyellow,minimum height=0.35cm,minimum width=0.35cm,draw,thick](out22) at ([xshift=0.1cm]or.east){\scriptsize \textbf {\textsl{2}}};  

    \node[embed,fill=white,minimum height=0.5cm,minimum width=0.5cm,draw,thick](add) at ([yshift=1cm]sm.north){};
    \node[textonly] (addput) at ([yshift=0.45cm]add.north) {}; 

    \draw[->,thick](emback.north)--(ln.south);
    \draw[->,thick](ln.north)--(sm.south);
    \draw[-,thick](add.north)--(add.south); 
    \draw[-,thick](add.east)--(add.west); 
    \draw[->,thick](add.north)--(addput.south);
    \draw[->,thick,rounded corners=3pt] ([yshift=0.2cm]ln.north) --([xshift=1.75cm,yshift=0.2cm]ln.north)-- ([xshift=1.75cm,yshift=2.06cm]ln.north)--(add.east);
    \draw[->,thick,rounded corners=3pt] (out22.east) --([yshift=0.45cm]sm.north)--(add.south);
    \draw[->,ultra thick,qblue]([xshift=-0.06cm]sm.west)--([xshift=-0.56cm]sm.west); 

    \node[background,draw,dotted,minimum height=2.4cm, minimum width=3.25cm] (comback) at ([yshift=1.78cm,xshift=0.11cm]add.north) {};
    \node[embed,minimum height=0.25cm,minimum width=0.25cm] (com1) at ([xshift=-1.42cm,yshift=0.97cm]comback.center) {};
    \node[textonly] (com1t) at ([xshift=0.96cm]com1.east) {Word embedding};

    \node[probr,fill=qblue,minimum height=0.2cm,minimum width=0.25cm] (com4) at ([yshift=-0.2cm]com1.south) {}; 
    \node[textonly] (com4t) at ([xshift=1.46cm]com4.east) {CTC Predicted Distribution};
    \node[probp,fill=qorange,minimum height=0.2cm,minimum width=0.25cm] (com5) at ([yshift=-0.2cm]com4.south) {}; 
    \node[textonly] (com5t) at ([xshift=1.21cm]com5.east) {Reference Distribution};
    
    \node[embed,fill=qblue,minimum height=0.25cm,minimum width=0.25cm] (com2) at ([yshift=-0.2cm]com5.south){};
    \node[textonly] (com2t1) at ([xshift=0.26cm]com2.east) {$\sum$(};
    \node[embed,minimum height=0.25cm,minimum width=0.25cm] (com2embd1) at ([xshift=0.1cm]com2t1.east) {};
    \node[textonly] (com2t2) at ([xshift=0.1cm]com2embd1.east) {$\times$};
    \node[probr,fill=qblue,minimum height=0.2cm,minimum width=0.25cm] (com2prob1) at ([xshift=0.1cm]com2t2.east) {}; 
    \node[textonly] (com2t3) at ([xshift=0.1cm]com2prob1.east) {)};
    
    \node[embed,fill=qorange,minimum height=0.25cm,minimum width=0.25cm] (com3) at ([yshift=-0.2cm]com2.south){};
    \node[textonly] (com3t1) at ([xshift=0.26cm]com3.east) {$\sum$(};
    \node[embed,minimum height=0.25cm,minimum width=0.25cm] (com3embd1) at ([xshift=0.1cm]com3t1.east) {};
    \node[textonly] (com3t2) at ([xshift=0.1cm]com3embd1.east) {$\times$};
    \node[probp,fill=qorange,minimum height=0.2cm,minimum width=0.25cm] (com3prob1) at ([xshift=0.1cm]com3t2.east) {}; 
    \node[textonly] (com3t3) at ([xshift=0.1cm]com3prob1.east) {)};
    \node[embed,fill=qyellow,minimum height=0.24cm,minimum width=0.24cm,draw,thick](out31) at ([yshift=-0.2cm]com3.south){\tiny \textbf {\textsl{1}} };
    \node[embed,fill=qyellow,minimum height=0.24cm,minimum width=0.24cm,draw,thick](out32) at ([yshift=-0.2cm]out31.south){\tiny \textbf {\textsl{2}}}; 
    \node[textonly,align=left] (com6t) at ([xshift=1.3cm]out31.east) {Predicted Representation};
    \node[textonly,align=left] (com7t) at ([xshift=1.15cm]out32.east) {Mixed Representation};

    \end{tikzpicture}
\end{minipage}
}
\caption{Overview of our CTC-NAST model. (a) The base architecture consisting of two CTC-guided encoders, (b): The cross-layer attention module, where the layer normalization is omitted for simplification, (c) Prediction-aware encoding, and its variant of curriculum learning mixing.}
\label{fig:fig1}
\end{figure*}

%% file: acl2023.bbl
\begin{thebibliography}{61}
\expandafter\ifx\csname natexlab\endcsname\relax\def\natexlab#1{#1}\fi

\bibitem[{Anastasopoulos and Chiang(2018)}]{Anastasopoulos_NAACL2018}
Antonios Anastasopoulos and David Chiang. 2018.
\newblock \href {https://doi.org/10.18653/v1/n18-1008} {Tied multitask learning
  for neural speech translation}.
\newblock In \emph{Proceedings of the 2018 Conference of the North American
  Chapter of the Association for Computational Linguistics: Human Language
  Technologies, {NAACL-HLT} 2018, New Orleans, Louisiana, USA, June 1-6, 2018,
  Volume 1 (Long Papers)}, pages 82--91. Association for Computational
  Linguistics.

\bibitem[{Bao et~al.(2022)Bao, Zhou, Huang, Wang, Qian, Dai, Chen, and
  Li}]{Bao_ACL2022}
Yu~Bao, Hao Zhou, Shujian Huang, Dongqi Wang, Lihua Qian, Xinyu Dai, Jiajun
  Chen, and Lei Li. 2022.
\newblock \href {https://doi.org/10.18653/v1/2022.acl-long.575} {latent-glat:
  Glancing at latent variables for parallel text generation}.
\newblock In \emph{Proceedings of the 60th Annual Meeting of the Association
  for Computational Linguistics (Volume 1: Long Papers), {ACL} 2022, Dublin,
  Ireland, May 22-27, 2022}, pages 8398--8409. Association for Computational
  Linguistics.

\bibitem[{Berard et~al.(2016)Berard, Pietquin, Servan, and
  Besacier}]{Berard_arxiv2016}
Alexandre Berard, Olivier Pietquin, Christophe Servan, and Laurent Besacier.
  2016.
\newblock \href {http://arxiv.org/abs/1612.01744} {Listen and translate: {A}
  proof of concept for end-to-end speech-to-text translation}.
\newblock \emph{CoRR}, abs/1612.01744.

\bibitem[{Chan et~al.(2020)Chan, Saharia, Hinton, Norouzi, and
  Jaitly}]{Chan_ICML2020}
William Chan, Chitwan Saharia, Geoffrey~E. Hinton, Mohammad Norouzi, and
  Navdeep Jaitly. 2020.
\newblock \href {http://proceedings.mlr.press/v119/chan20b.html} {Imputer:
  Sequence modelling via imputation and dynamic programming}.
\newblock In \emph{Proceedings of the 37th International Conference on Machine
  Learning, {ICML} 2020, 13-18 July 2020, Virtual Event}, volume 119 of
  \emph{Proceedings of Machine Learning Research}, pages 1403--1413. {PMLR}.

\bibitem[{Cheng et~al.(2022)Cheng, Dong, Yue, Ko, Wang, and
  Zou}]{Cheng_Corr2022}
Xuxin Cheng, Qianqian Dong, Fengpeng Yue, Tom Ko, Mingxuan Wang, and Yuexian
  Zou. 2022.
\newblock \href {https://doi.org/10.48550/arXiv.2212.03657} {{M3ST:} mix at
  three levels for speech translation}.
\newblock \emph{CoRR}, abs/2212.03657.

\bibitem[{Chuang et~al.(2021)Chuang, Chuang, Chang, and Lee}]{Chuang_ACL2021}
Shun{-}Po Chuang, Yung{-}Sung Chuang, Chih{-}Chiang Chang, and Hung{-}yi Lee.
  2021.
\newblock \href {https://doi.org/10.18653/v1/2021.findings-acl.92}
  {Investigating the reordering capability in ctc-based non-autoregressive
  end-to-end speech translation}.
\newblock In \emph{Findings of the Association for Computational Linguistics:
  {ACL/IJCNLP} 2021, Online Event, August 1-6, 2021}, volume {ACL/IJCNLP} 2021
  of \emph{Findings of {ACL}}, pages 1068--1077. Association for Computational
  Linguistics.

\bibitem[{Ding et~al.(2021)Ding, Wang, Liu, Wong, Tao, and Tu}]{Ding_ACL2020}
Liang Ding, Longyue Wang, Xuebo Liu, Derek~F. Wong, Dacheng Tao, and Zhaopeng
  Tu. 2021.
\newblock \href {https://doi.org/10.18653/v1/2021.acl-long.266} {Rejuvenating
  low-frequency words: Making the most of parallel data in non-autoregressive
  translation}.
\newblock In \emph{Proceedings of the 59th Annual Meeting of the Association
  for Computational Linguistics and the 11th International Joint Conference on
  Natural Language Processing, {ACL/IJCNLP} 2021, (Volume 1: Long Papers),
  Virtual Event, August 1-6, 2021}, pages 3431--3441. Association for
  Computational Linguistics.

\bibitem[{Duong et~al.(2016)Duong, Anastasopoulos, Chiang, Bird, and
  Cohn}]{Duong_naacl2016}
Long Duong, Antonios Anastasopoulos, David Chiang, Steven Bird, and Trevor
  Cohn. 2016.
\newblock \href {https://doi.org/10.18653/v1/n16-1109} {An attentional model
  for speech translation without transcription}.
\newblock In \emph{{NAACL} {HLT} 2016, The 2016 Conference of the North
  American Chapter of the Association for Computational Linguistics: Human
  Language Technologies, San Diego California, USA, June 12-17, 2016}, pages
  949--959. The Association for Computational Linguistics.

\bibitem[{Fang et~al.(2022)Fang, Ye, Li, Feng, and Wang}]{Fang_ACL2022}
Qingkai Fang, Rong Ye, Lei Li, Yang Feng, and Mingxuan Wang. 2022.
\newblock \href {https://doi.org/10.18653/v1/2022.acl-long.486} {{STEMM:}
  self-learning with speech-text manifold mixup for speech translation}.
\newblock In \emph{Proceedings of the 60th Annual Meeting of the Association
  for Computational Linguistics (Volume 1: Long Papers), {ACL} 2022, Dublin,
  Ireland, May 22-27, 2022}, pages 7050--7062. Association for Computational
  Linguistics.

\bibitem[{Gangi et~al.(2019)Gangi, Cattoni, Bentivogli, Negri, and
  Turchi}]{GangiCBNT19}
Mattia Antonino~Di Gangi, Roldano Cattoni, Luisa Bentivogli, Matteo Negri, and
  Marco Turchi. 2019.
\newblock \href {https://doi.org/10.18653/v1/n19-1202} {Must-c: a multilingual
  speech translation corpus}.
\newblock In \emph{Proceedings of the 2019 Conference of the North American
  Chapter of the Association for Computational Linguistics: Human Language
  Technologies, {NAACL-HLT} 2019, Minneapolis, MN, USA, June 2-7, 2019, Volume
  1 (Long and Short Papers)}, pages 2012--2017. Association for Computational
  Linguistics.

\bibitem[{Ghazvininejad et~al.(2019)Ghazvininejad, Levy, Liu, and
  Zettlemoyer}]{Ghazvininejad_EMNLP2019}
Marjan Ghazvininejad, Omer Levy, Yinhan Liu, and Luke Zettlemoyer. 2019.
\newblock \href {https://doi.org/10.18653/v1/D19-1633} {Mask-predict: Parallel
  decoding of conditional masked language models}.
\newblock In \emph{Proceedings of the 2019 Conference on Empirical Methods in
  Natural Language Processing and the 9th International Joint Conference on
  Natural Language Processing, {EMNLP-IJCNLP} 2019, Hong Kong, China, November
  3-7, 2019}, pages 6111--6120. Association for Computational Linguistics.

\bibitem[{Graves et~al.(2006)Graves, Fern{\'{a}}ndez, Gomez, and
  Schmidhuber}]{Graves_ACL2006}
Alex Graves, Santiago Fern{\'{a}}ndez, Faustino~J. Gomez, and J{\"{u}}rgen
  Schmidhuber. 2006.
\newblock \href {https://doi.org/10.1145/1143844.1143891} {Connectionist
  temporal classification: labelling unsegmented sequence data with recurrent
  neural networks}.
\newblock In \emph{Machine Learning, Proceedings of the Twenty-Third
  International Conference {(ICML} 2006), Pittsburgh, Pennsylvania, USA, June
  25-29, 2006}, volume 148 of \emph{{ACM} International Conference Proceeding
  Series}, pages 369--376. {ACM}.

\bibitem[{Gu et~al.(2018)Gu, Bradbury, Xiong, Li, and Socher}]{Gu_ICLR2018}
Jiatao Gu, James Bradbury, Caiming Xiong, Victor O.~K. Li, and Richard Socher.
  2018.
\newblock \href {https://openreview.net/forum?id=B1l8BtlCb} {Non-autoregressive
  neural machine translation}.
\newblock In \emph{6th International Conference on Learning Representations,
  {ICLR} 2018, Vancouver, BC, Canada, April 30 - May 3, 2018, Conference Track
  Proceedings}. OpenReview.net.

\bibitem[{Gu and Kong(2021)}]{Gu_ACL2021}
Jiatao Gu and Xiang Kong. 2021.
\newblock \href {https://doi.org/10.18653/v1/2021.findings-acl.11} {Fully
  non-autoregressive neural machine translation: Tricks of the trade}.
\newblock In \emph{Findings of the Association for Computational Linguistics:
  {ACL/IJCNLP} 2021, Online Event, August 1-6, 2021}, volume {ACL/IJCNLP} 2021
  of \emph{Findings of {ACL}}, pages 120--133. Association for Computational
  Linguistics.

\bibitem[{Guo et~al.(2019)Guo, Tan, He, Qin, Xu, and Liu}]{Guo_AAAI2019}
Junliang Guo, Xu~Tan, Di~He, Tao Qin, Linli Xu, and Tie{-}Yan Liu. 2019.
\newblock \href {https://doi.org/10.1609/aaai.v33i01.33013723}
  {Non-autoregressive neural machine translation with enhanced decoder input}.
\newblock In \emph{The Thirty-Third {AAAI} Conference on Artificial
  Intelligence, {AAAI} 2019, The Thirty-First Innovative Applications of
  Artificial Intelligence Conference, {IAAI} 2019, The Ninth {AAAI} Symposium
  on Educational Advances in Artificial Intelligence, {EAAI} 2019, Honolulu,
  Hawaii, USA, January 27 - February 1, 2019}, pages 3723--3730. {AAAI} Press.

\bibitem[{Hannun(2017)}]{hannun2017distill}
Awni Hannun. 2017.
\newblock \href {https://doi.org/10.23915/distill.00008} {Sequence modeling
  with ctc}.
\newblock \emph{Distill}.
\newblock Https://distill.pub/2017/ctc.

\bibitem[{Higuchi et~al.(2021{\natexlab{a}})Higuchi, Chen, Fujita, Inaguma,
  Komatsu, Lee, Nozaki, Wang, and Watanabe}]{Higuchi_ASRU2021}
Yosuke Higuchi, Nanxin Chen, Yuya Fujita, Hirofumi Inaguma, Tatsuya Komatsu,
  Jaesong Lee, Jumon Nozaki, Tianzi Wang, and Shinji Watanabe.
  2021{\natexlab{a}}.
\newblock \href {https://doi.org/10.1109/ASRU51503.2021.9688157} {A comparative
  study on non-autoregressive modelings for speech-to-text generation}.
\newblock In \emph{{IEEE} Automatic Speech Recognition and Understanding
  Workshop, {ASRU} 2021, Cartagena, Colombia, December 13-17, 2021}, pages
  47--54. {IEEE}.

\bibitem[{Higuchi et~al.(2021{\natexlab{b}})Higuchi, Inaguma, Watanabe, Ogawa,
  and Kobayashi}]{Higuchi_ICASSP2021}
Yosuke Higuchi, Hirofumi Inaguma, Shinji Watanabe, Tetsuji Ogawa, and Tetsunori
  Kobayashi. 2021{\natexlab{b}}.
\newblock \href {https://doi.org/10.1109/ICASSP39728.2021.9414198} {Improved
  mask-ctc for non-autoregressive end-to-end asr}.
\newblock In \emph{ICASSP 2021 - 2021 IEEE International Conference on
  Acoustics, Speech and Signal Processing (ICASSP)}, pages 8363--8367.

\bibitem[{Higuchi et~al.(2020)Higuchi, Watanabe, Chen, Ogawa, and
  Kobayashi}]{higuchi20b_interspeech}
Yosuke Higuchi, Shinji Watanabe, Nanxin Chen, Tetsuji Ogawa, and Tetsunori
  Kobayashi. 2020.
\newblock \href {https://doi.org/10.21437/Interspeech.2020-2404} {{Mask CTC:
  Non-Autoregressive End-to-End ASR with CTC and Mask Predict}}.
\newblock In \emph{Proc. Interspeech 2020}, pages 3655--3659.

\bibitem[{Huang et~al.(2022)Huang, Zhou, Za{\"{\i}}ane, Mou, and
  Li}]{Huang_AAAI2022}
Chenyang Huang, Hao Zhou, Osmar~R. Za{\"{\i}}ane, Lili Mou, and Lei Li. 2022.
\newblock \href {https://ojs.aaai.org/index.php/AAAI/article/view/21323}
  {Non-autoregressive translation with layer-wise prediction and deep
  supervision}.
\newblock In \emph{Thirty-Sixth {AAAI} Conference on Artificial Intelligence,
  {AAAI} 2022, Thirty-Fourth Conference on Innovative Applications of
  Artificial Intelligence, {IAAI} 2022, The Twelveth Symposium on Educational
  Advances in Artificial Intelligence, {EAAI} 2022 Virtual Event, February 22 -
  March 1, 2022}, pages 10776--10784. {AAAI} Press.

\bibitem[{Inaguma et~al.(2021{\natexlab{a}})Inaguma, Higuchi, Duh, Kawahara,
  and Watanabe}]{Inaguma_CORR2021}
Hirofumi Inaguma, Yosuke Higuchi, Kevin Duh, Tatsuya Kawahara, and Shinji
  Watanabe. 2021{\natexlab{a}}.
\newblock \href {http://arxiv.org/abs/2109.04411} {Non-autoregressive
  end-to-end speech translation with parallel autoregressive rescoring}.
\newblock \emph{CoRR}, abs/2109.04411.

\bibitem[{Inaguma et~al.(2021{\natexlab{b}})Inaguma, Higuchi, Duh, Kawahara,
  and Watanabe}]{Inaguma_ICASSP2021}
Hirofumi Inaguma, Yosuke Higuchi, Kevin Duh, Tatsuya Kawahara, and Shinji
  Watanabe. 2021{\natexlab{b}}.
\newblock \href {https://doi.org/10.1109/ICASSP39728.2021.9415093} {{ORTHROS:}
  non-autoregressive end-to-end speech translation with dual-decoder}.
\newblock In \emph{{IEEE} International Conference on Acoustics, Speech and
  Signal Processing, {ICASSP} 2021, Toronto, ON, Canada, June 6-11, 2021},
  pages 7503--7507. {IEEE}.

\bibitem[{Inaguma et~al.(2020)Inaguma, Kiyono, Duh, Karita, Yalta, Hayashi, and
  Watanabe}]{Inaguma_ACL2020}
Hirofumi Inaguma, Shun Kiyono, Kevin Duh, Shigeki Karita, Nelson Yalta, Tomoki
  Hayashi, and Shinji Watanabe. 2020.
\newblock \href {https://doi.org/10.18653/v1/2020.acl-demos.34} {Espnet-st:
  All-in-one speech translation toolkit}.
\newblock In \emph{Proceedings of the 58th Annual Meeting of the Association
  for Computational Linguistics: System Demonstrations, {ACL} 2020, Online,
  July 5-10, 2020}, pages 302--311. Association for Computational Linguistics.

\bibitem[{Karita et~al.(2019)Karita, Soplin, Watanabe, Delcroix, Ogawa, and
  Nakatani}]{Karita_ISCA2019}
Shigeki Karita, Nelson Enrique~Yalta Soplin, Shinji Watanabe, Marc Delcroix,
  Atsunori Ogawa, and Tomohiro Nakatani. 2019.
\newblock \href {https://doi.org/10.21437/Interspeech.2019-1938} {Improving
  transformer-based end-to-end speech recognition with connectionist temporal
  classification and language model integration}.
\newblock In \emph{Interspeech 2019, 20th Annual Conference of the
  International Speech Communication Association, Graz, Austria, 15-19
  September 2019}, pages 1408--1412. {ISCA}.

\bibitem[{Kasai et~al.(2020)Kasai, Cross, Ghazvininejad, and
  Gu}]{Kasai_ICML2020}
Jungo Kasai, James Cross, Marjan Ghazvininejad, and Jiatao Gu. 2020.
\newblock \href {http://proceedings.mlr.press/v119/kasai20a.html}
  {Non-autoregressive machine translation with disentangled context
  transformer}.
\newblock In \emph{Proceedings of the 37th International Conference on Machine
  Learning, {ICML} 2020, 13-18 July 2020, Virtual Event}, volume 119 of
  \emph{Proceedings of Machine Learning Research}, pages 5144--5155. {PMLR}.

\bibitem[{Kim et~al.(2022)Kim, Gholami, Shaw, Lee, Mangalam, Malik, Mahoney,
  and Keutzer}]{Kim_Corr2022}
Sehoon Kim, Amir Gholami, Albert~E. Shaw, Nicholas Lee, Karttikeya Mangalam,
  Jitendra Malik, Michael~W. Mahoney, and Kurt Keutzer. 2022.
\newblock \href {https://doi.org/10.48550/arXiv.2206.00888} {Squeezeformer: An
  efficient transformer for automatic speech recognition}.
\newblock \emph{CoRR}, abs/2206.00888.

\bibitem[{Kim and Rush(2016)}]{Kim_EMNLP2016}
Yoon Kim and Alexander~M. Rush. 2016.
\newblock \href {https://doi.org/10.18653/v1/d16-1139} {Sequence-level
  knowledge distillation}.
\newblock In \emph{Proceedings of the 2016 Conference on Empirical Methods in
  Natural Language Processing, {EMNLP} 2016, Austin, Texas, USA, November 1-4,
  2016}, pages 1317--1327. The Association for Computational Linguistics.

\bibitem[{Koehn et~al.(2007)Koehn, Hoang, Birch, Callison{-}Burch, Federico,
  Bertoldi, Cowan, Shen, Moran, Zens, Dyer, Bojar, Constantin, and
  Herbst}]{Koehn_ACL2007}
Philipp Koehn, Hieu Hoang, Alexandra Birch, Chris Callison{-}Burch, Marcello
  Federico, Nicola Bertoldi, Brooke Cowan, Wade Shen, Christine Moran, Richard
  Zens, Chris Dyer, Ondrej Bojar, Alexandra Constantin, and Evan Herbst. 2007.
\newblock \href {https://www.aclweb.org/anthology/P07-2045/} {Moses: Open
  source toolkit for statistical machine translation}.
\newblock In \emph{{ACL} 2007, Proceedings of the 45th Annual Meeting of the
  Association for Computational Linguistics, June 23-30, 2007, Prague, Czech
  Republic}. The Association for Computational Linguistics.

\bibitem[{Lee and Watanabe(2021)}]{Lee_ICASSP2021}
Jaesong Lee and Shinji Watanabe. 2021.
\newblock \href {https://doi.org/10.1109/ICASSP39728.2021.9414594}
  {Intermediate loss regularization for ctc-based speech recognition}.
\newblock In \emph{{IEEE} International Conference on Acoustics, Speech and
  Signal Processing, {ICASSP} 2021, Toronto, ON, Canada, June 6-11, 2021},
  pages 6224--6228. {IEEE}.

\bibitem[{Lee et~al.(2018)Lee, Mansimov, and Cho}]{Lee_EMNLP2018}
Jason Lee, Elman Mansimov, and Kyunghyun Cho. 2018.
\newblock \href {https://doi.org/10.18653/v1/d18-1149} {Deterministic
  non-autoregressive neural sequence modeling by iterative refinement}.
\newblock In \emph{Proceedings of the 2018 Conference on Empirical Methods in
  Natural Language Processing, Brussels, Belgium, October 31 - November 4,
  2018}, pages 1173--1182. Association for Computational Linguistics.

\bibitem[{Libovick{\'{y}} and Helcl(2018)}]{Libovick_EMNLP2018}
Jindrich Libovick{\'{y}} and Jindrich Helcl. 2018.
\newblock \href {https://doi.org/10.18653/v1/d18-1336} {End-to-end
  non-autoregressive neural machine translation with connectionist temporal
  classification}.
\newblock In \emph{Proceedings of the 2018 Conference on Empirical Methods in
  Natural Language Processing, Brussels, Belgium, October 31 - November 4,
  2018}, pages 3016--3021. Association for Computational Linguistics.

\bibitem[{Liu et~al.(2020)Liu, Zhu, Zhang, and Zong}]{Liu_corr2020}
Yuchen Liu, Junnan Zhu, Jiajun Zhang, and Chengqing Zong. 2020.
\newblock \href {http://arxiv.org/abs/2010.14920} {Bridging the modality gap
  for speech-to-text translation}.
\newblock \emph{CoRR}, abs/2010.14920.

\bibitem[{Mathias and Byrne(2006)}]{Mathias_ICASSP2006}
Lambert Mathias and William Byrne. 2006.
\newblock \href {https://doi.org/10.1109/ICASSP.2006.1660082} {Statistical
  phrase-based speech translation}.
\newblock In \emph{2006 {IEEE} International Conference on Acoustics Speech and
  Signal Processing, {ICASSP} 2006, Toulouse, France, May 14-19, 2006}, pages
  561--564. {IEEE}.

\bibitem[{Ney(1999)}]{Ney_IEEE1999}
Hermann Ney. 1999.
\newblock \href {https://doi.org/10.1109/ICASSP.1999.758176} {Speech
  translation: coupling of recognition and translation}.
\newblock In \emph{Proceedings of the 1999 {IEEE} International Conference on
  Acoustics, Speech, and Signal Processing, {ICASSP} '99, Phoenix, Arizona,
  USA, March 15-19, 1999}, pages 517--520. {IEEE} Computer Society.

\bibitem[{Nozaki and Komatsu(2021)}]{Nozaki_ISCA2021}
Jumon Nozaki and Tatsuya Komatsu. 2021.
\newblock \href {https://doi.org/10.21437/Interspeech.2021-911} {Relaxing the
  conditional independence assumption of ctc-based {ASR} by conditioning on
  intermediate predictions}.
\newblock In \emph{Interspeech 2021, 22nd Annual Conference of the
  International Speech Communication Association, Brno, Czechia, 30 August - 3
  September 2021}, pages 3735--3739. {ISCA}.

\bibitem[{Ott et~al.(2019)Ott, Edunov, Baevski, Fan, Gross, Ng, Grangier, and
  Auli}]{Ott_NAACL2019}
Myle Ott, Sergey Edunov, Alexei Baevski, Angela Fan, Sam Gross, Nathan Ng,
  David Grangier, and Michael Auli. 2019.
\newblock \href {https://doi.org/10.18653/v1/n19-4009} {fairseq: {A} fast,
  extensible toolkit for sequence modeling}.
\newblock In \emph{Proceedings of the 2019 Conference of the North American
  Chapter of the Association for Computational Linguistics: Human Language
  Technologies, {NAACL-HLT} 2019, Minneapolis, MN, USA, June 2-7, 2019,
  Demonstrations}, pages 48--53. Association for Computational Linguistics.

\bibitem[{Post(2018)}]{Post_wmt18}
Matt Post. 2018.
\newblock \href {https://doi.org/10.18653/v1/w18-6319} {A call for clarity in
  reporting {BLEU} scores}.
\newblock In \emph{Proceedings of the Third Conference on Machine Translation:
  Research Papers, {WMT} 2018, Belgium, Brussels, October 31 - November 1,
  2018}, pages 186--191. Association for Computational Linguistics.

\bibitem[{Post et~al.(2013)Post, Kumar, Lopez, Karakos, Callison{-}Burch, and
  Khudanpur}]{Post_IWSLT2013}
Matt Post, Gaurav Kumar, Adam Lopez, Damianos~G. Karakos, Chris
  Callison{-}Burch, and Sanjeev Khudanpur. 2013.
\newblock \href {https://aclanthology.org/2013.iwslt-papers.14} {Improved
  speech-to-text translation with the fisher and callhome spanish-english
  speech translation corpus}.
\newblock In \emph{Proceedings of the 10th International Workshop on Spoken
  Language Translation: Papers, Heidelberg, Germany, December 5-6, 2013}.

\bibitem[{Qian et~al.(2021)Qian, Zhou, Bao, Wang, Qiu, Zhang, Yu, and
  Li}]{Qian_ACL2021}
Lihua Qian, Hao Zhou, Yu~Bao, Mingxuan Wang, Lin Qiu, Weinan Zhang, Yong Yu,
  and Lei Li. 2021.
\newblock \href {https://doi.org/10.18653/v1/2021.acl-long.155} {Glancing
  transformer for non-autoregressive neural machine translation}.
\newblock In \emph{Proceedings of the 59th Annual Meeting of the Association
  for Computational Linguistics and the 11th International Joint Conference on
  Natural Language Processing, {ACL/IJCNLP} 2021, (Volume 1: Long Papers),
  Virtual Event, August 1-6, 2021}, pages 1993--2003. Association for
  Computational Linguistics.

\bibitem[{Ran et~al.(2020)Ran, Lin, Li, and Zhou}]{Ran_ACL2020}
Qiu Ran, Yankai Lin, Peng Li, and Jie Zhou. 2020.
\newblock \href {https://doi.org/10.18653/v1/2020.acl-main.277} {Learning to
  recover from multi-modality errors for non-autoregressive neural machine
  translation}.
\newblock In \emph{Proceedings of the 58th Annual Meeting of the Association
  for Computational Linguistics, {ACL} 2020, Online, July 5-10, 2020}, pages
  3059--3069. Association for Computational Linguistics.

\bibitem[{Ran et~al.(2021)Ran, Lin, Li, and Zhou}]{Ran_AAAI2021}
Qiu Ran, Yankai Lin, Peng Li, and Jie Zhou. 2021.
\newblock \href {https://ojs.aaai.org/index.php/AAAI/article/view/17618}
  {Guiding non-autoregressive neural machine translation decoding with
  reordering information}.
\newblock In \emph{Thirty-Fifth {AAAI} Conference on Artificial Intelligence,
  {AAAI} 2021, Thirty-Third Conference on Innovative Applications of Artificial
  Intelligence, {IAAI} 2021, The Eleventh Symposium on Educational Advances in
  Artificial Intelligence, {EAAI} 2021, Virtual Event, February 2-9, 2021},
  pages 13727--13735. {AAAI} Press.

\bibitem[{Saharia et~al.(2020)Saharia, Chan, Saxena, and
  Norouzi}]{Saharia_EMNLP2020}
Chitwan Saharia, William Chan, Saurabh Saxena, and Mohammad Norouzi. 2020.
\newblock \href {https://doi.org/10.18653/v1/2020.emnlp-main.83}
  {Non-autoregressive machine translation with latent alignments}.
\newblock In \emph{Proceedings of the 2020 Conference on Empirical Methods in
  Natural Language Processing, {EMNLP} 2020, Online, November 16-20, 2020},
  pages 1098--1108. Association for Computational Linguistics.

\bibitem[{Shu et~al.(2020)Shu, Lee, Nakayama, and Cho}]{Shu_AAAI2020}
Raphael Shu, Jason Lee, Hideki Nakayama, and Kyunghyun Cho. 2020.
\newblock \href {https://ojs.aaai.org/index.php/AAAI/article/view/6413}
  {Latent-variable non-autoregressive neural machine translation with
  deterministic inference using a delta posterior}.
\newblock In \emph{The Thirty-Fourth {AAAI} Conference on Artificial
  Intelligence, {AAAI} 2020, The Thirty-Second Innovative Applications of
  Artificial Intelligence Conference, {IAAI} 2020, The Tenth {AAAI} Symposium
  on Educational Advances in Artificial Intelligence, {EAAI} 2020, New York,
  NY, USA, February 7-12, 2020}, pages 8846--8853. {AAAI} Press.

\bibitem[{Song et~al.(2021)Song, Kim, and Yoon}]{Song_EMNLP2021}
Jongyoon Song, Sungwon Kim, and Sungroh Yoon. 2021.
\newblock \href {https://doi.org/10.18653/v1/2021.emnlp-main.1} {Alignart:
  Non-autoregressive neural machine translation by jointly learning to estimate
  alignment and translate}.
\newblock In \emph{Proceedings of the 2021 Conference on Empirical Methods in
  Natural Language Processing, {EMNLP} 2021, Virtual Event / Punta Cana,
  Dominican Republic, 7-11 November, 2021}, pages 1--14. Association for
  Computational Linguistics.

\bibitem[{Stern et~al.(2019)Stern, Chan, Kiros, and Uszkoreit}]{Stern_ICML2019}
Mitchell Stern, William Chan, Jamie Kiros, and Jakob Uszkoreit. 2019.
\newblock \href {http://proceedings.mlr.press/v97/stern19a.html} {Insertion
  transformer: Flexible sequence generation via insertion operations}.
\newblock In \emph{Proceedings of the 36th International Conference on Machine
  Learning, {ICML} 2019, 9-15 June 2019, Long Beach, California, {USA}},
  volume~97 of \emph{Proceedings of Machine Learning Research}, pages
  5976--5985. {PMLR}.

\bibitem[{Wang et~al.(2020{\natexlab{a}})Wang, Tang, Ma, Wu, Okhonko, and
  Pino}]{Wang_AACL2020}
Changhan Wang, Yun Tang, Xutai Ma, Anne Wu, Dmytro Okhonko, and Juan~Miguel
  Pino. 2020{\natexlab{a}}.
\newblock \href {https://aclanthology.org/2020.aacl-demo.6/} {Fairseq {S2T:}
  fast speech-to-text modeling with fairseq}.
\newblock In \emph{Proceedings of the 1st Conference of the Asia-Pacific
  Chapter of the Association for Computational Linguistics and the 10th
  International Joint Conference on Natural Language Processing: System
  Demonstrations, {AACL/IJCNLP} 2020, Suzhou, China, December 4-7, 2020}, pages
  33--39. Association for Computational Linguistics.

\bibitem[{Wang et~al.(2020{\natexlab{b}})Wang, Wu, Liu, Yang, and
  Zhou}]{Wang_aaai2020}
Chengyi Wang, Yu~Wu, Shujie Liu, Zhenglu Yang, and Ming Zhou.
  2020{\natexlab{b}}.
\newblock \href {https://aaai.org/ojs/index.php/AAAI/article/view/6452}
  {Bridging the gap between pre-training and fine-tuning for end-to-end speech
  translation}.
\newblock In \emph{The Thirty-Fourth {AAAI} Conference on Artificial
  Intelligence, {AAAI} 2020, The Thirty-Second Innovative Applications of
  Artificial Intelligence Conference, {IAAI} 2020, The Tenth {AAAI} Symposium
  on Educational Advances in Artificial Intelligence, {EAAI} 2020, New York,
  NY, USA, February 7-12, 2020}, pages 9161--9168. {AAAI} Press.

\bibitem[{Wang et~al.(2020{\natexlab{c}})Wang, Wu, Liu, Zhou, and
  Yang}]{Wang_acl2020}
Chengyi Wang, Yu~Wu, Shujie Liu, Ming Zhou, and Zhenglu Yang.
  2020{\natexlab{c}}.
\newblock \href {https://doi.org/10.18653/v1/2020.acl-main.344} {Curriculum
  pre-training for end-to-end speech translation}.
\newblock In \emph{Proceedings of the 58th Annual Meeting of the Association
  for Computational Linguistics, {ACL} 2020, Online, July 5-10, 2020}, pages
  3728--3738. Association for Computational Linguistics.

\bibitem[{Wang et~al.(2019)Wang, Tian, He, Qin, Zhai, and Liu}]{Wang_AAAI2019}
Yiren Wang, Fei Tian, Di~He, Tao Qin, ChengXiang Zhai, and Tie{-}Yan Liu. 2019.
\newblock \href {https://doi.org/10.1609/aaai.v33i01.33015377}
  {Non-autoregressive machine translation with auxiliary regularization}.
\newblock In \emph{The Thirty-Third {AAAI} Conference on Artificial
  Intelligence, {AAAI} 2019, The Thirty-First Innovative Applications of
  Artificial Intelligence Conference, {IAAI} 2019, The Ninth {AAAI} Symposium
  on Educational Advances in Artificial Intelligence, {EAAI} 2019, Honolulu,
  Hawaii, USA, January 27 - February 1, 2019}, pages 5377--5384. {AAAI} Press.

\bibitem[{Watanabe et~al.(2017)Watanabe, Hori, Kim, Hershey, and
  Hayashi}]{Watanabe_IEEE2017}
Shinji Watanabe, Takaaki Hori, Suyoun Kim, John~R. Hershey, and Tomoki Hayashi.
  2017.
\newblock \href {https://doi.org/10.1109/JSTSP.2017.2763455} {Hybrid
  ctc/attention architecture for end-to-end speech recognition}.
\newblock \emph{{IEEE} J. Sel. Top. Signal Process.}, 11(8):1240--1253.

\bibitem[{Weiss et~al.(2017)Weiss, Chorowski, Jaitly, Wu, and
  Chen}]{Weiss_ISCA2017}
Ron~J. Weiss, Jan Chorowski, Navdeep Jaitly, Yonghui Wu, and Zhifeng Chen.
  2017.
\newblock \href
  {http://www.isca-speech.org/archive/Interspeech\_2017/abstracts/0503.html}
  {Sequence-to-sequence models can directly translate foreign speech}.
\newblock In \emph{Interspeech 2017, 18th Annual Conference of the
  International Speech Communication Association, Stockholm, Sweden, August
  20-24, 2017}, pages 2625--2629. {ISCA}.

\bibitem[{Xu et~al.(2021)Xu, Hu, Li, Zhang, Huang, Ju, Xiao, and
  Zhu}]{Xu_ACL2021}
Chen Xu, Bojie Hu, Yanyang Li, Yuhao Zhang, Shen Huang, Qi~Ju, Tong Xiao, and
  Jingbo Zhu. 2021.
\newblock \href {https://doi.org/10.18653/v1/2021.acl-long.204} {Stacked
  acoustic-and-textual encoding: Integrating the pre-trained models into speech
  translation encoders}.
\newblock In \emph{Proceedings of the 59th Annual Meeting of the Association
  for Computational Linguistics and the 11th International Joint Conference on
  Natural Language Processing, {ACL/IJCNLP} 2021, (Volume 1: Long Papers),
  Virtual Event, August 1-6, 2021}, pages 2619--2630. Association for
  Computational Linguistics.

\bibitem[{Xu et~al.(2023)Xu, Zhang, Jiao, Liu, Hu, Zeng, Xiao, Ma, Wang, and
  Zhu}]{xu2023pds}
Chen Xu, Yuhao Zhang, Chengbo Jiao, Xiaoqian Liu, Chi Hu, Xin Zeng, Tong Xiao,
  Anxiang Ma, Huizhen Wang, and Jingbo Zhu. 2023.
\newblock Bridging the granularity gap for acoustic modeling.
\newblock In \emph{Findings of the Association for Computational Linguistics:
  ACL 2023}. Association for Computational Linguistics.

\bibitem[{Yan et~al.(2022)Yan, Dalmia, Higuchi, Neubig, Metze, Black, and
  Watanabe}]{Yan_Corr2022}
Brian Yan, Siddharth Dalmia, Yosuke Higuchi, Graham Neubig, Florian Metze,
  Alan~W. Black, and Shinji Watanabe. 2022.
\newblock \href {https://doi.org/10.48550/arXiv.2210.05200} {{CTC} alignments
  improve autoregressive translation}.
\newblock \emph{CoRR}, abs/2210.05200.

\bibitem[{Yang et~al.(2018)Yang, Tu, Wong, Meng, Chao, and
  Zhang}]{Yang_emnlp2018}
Baosong Yang, Zhaopeng Tu, Derek~F. Wong, Fandong Meng, Lidia~S. Chao, and Tong
  Zhang. 2018.
\newblock \href {https://doi.org/10.18653/v1/d18-1475} {Modeling localness for
  self-attention networks}.
\newblock In \emph{Proceedings of the 2018 Conference on Empirical Methods in
  Natural Language Processing, Brussels, Belgium, October 31 - November 4,
  2018}, pages 4449--4458. Association for Computational Linguistics.

\bibitem[{Ye et~al.(2021)Ye, Wang, and Li}]{Ye_interspeech2021}
Rong Ye, Mingxuan Wang, and Lei Li. 2021.
\newblock \href {https://doi.org/10.21437/Interspeech.2021-1065} {End-to-end
  speech translation via cross-modal progressive training}.
\newblock In \emph{Interspeech 2021, 22nd Annual Conference of the
  International Speech Communication Association, Brno, Czechia, 30 August - 3
  September 2021}, pages 2267--2271. {ISCA}.

\bibitem[{Ye et~al.(2022)Ye, Wang, and Li}]{Ye_NAACL2022}
Rong Ye, Mingxuan Wang, and Lei Li. 2022.
\newblock \href {https://doi.org/10.18653/v1/2022.naacl-main.376} {Cross-modal
  contrastive learning for speech translation}.
\newblock In \emph{Proceedings of the 2022 Conference of the North American
  Chapter of the Association for Computational Linguistics: Human Language
  Technologies, {NAACL} 2022, Seattle, WA, United States, July 10-15, 2022},
  pages 5099--5113. Association for Computational Linguistics.

\bibitem[{Zhang et~al.(2022{\natexlab{a}})Zhang, Haddow, and
  Sennrich}]{Zhang_corr2022}
Biao Zhang, Barry Haddow, and Rico Sennrich. 2022{\natexlab{a}}.
\newblock \href {https://doi.org/10.48550/arXiv.2206.04571} {Revisiting
  end-to-end speech-to-text translation from scratch}.
\newblock \emph{CoRR}, abs/2206.04571.

\bibitem[{Zhang et~al.(2022{\natexlab{b}})Zhang, Xu, Hu, Zhang, Xiao, and
  Zhu}]{zhang2022improving}
Yuhao Zhang, Chen Xu, Bojie Hu, Chunliang Zhang, Tong Xiao, and Jingbo Zhu.
  2022{\natexlab{b}}.
\newblock \href {https://doi.org/10.48550/arXiv.2212.01778} {Improving
  end-to-end speech translation by leveraging auxiliary speech and text data}.
\newblock \emph{CoRR}, abs/2212.01778.

\bibitem[{Zhao et~al.(2021)Zhao, Wang, Dong, Ye, and Li}]{Zhao_ACL2021}
Chengqi Zhao, Mingxuan Wang, Qianqian Dong, Rong Ye, and Lei Li. 2021.
\newblock \href {https://aclanthology.org/2021.acl-demo.7} {Neurst: Neural
  speech translation toolkit}.
\newblock In \emph{Proceedings of the Joint Conference of the 59th Annual
  Meeting of the Association for Computational Linguistics and the 11th
  International Joint Conference on Natural Language Processing, {ACL} 2021 -
  System Demonstrations, Online, August 1-6, 2021}, pages 55--62. Association
  for Computational Linguistics.

\bibitem[{Zhou and Keung(2020)}]{Zhou_ACL2020}
Jiawei Zhou and Phillip Keung. 2020.
\newblock \href {https://doi.org/10.18653/v1/2020.acl-main.171} {Improving
  non-autoregressive neural machine translation with monolingual data}.
\newblock In \emph{Proceedings of the 58th Annual Meeting of the Association
  for Computational Linguistics, {ACL} 2020, Online, July 5-10, 2020}, pages
  1893--1898. Association for Computational Linguistics.

\end{thebibliography}
